\documentclass[aoas,preprint]{imsart}

\usepackage{amsthm,amsmath,natbib}
\usepackage{epsfig,amssymb,latexsym,graphicx,algorithm}
\usepackage{array,enumitem,url,booktabs}
\RequirePackage[OT1]{fontenc}
\RequirePackage[colorlinks,citecolor=blue,urlcolor=blue]{hyperref}
\arxiv{arXiv:1502.07190}

\startlocaldefs
\numberwithin{equation}{section}

\newtheorem{results}{Results}
\newcommand*{\Cdot}{\raisebox{-0.25ex}{\scalebox{1.2}{$\cdot$}}}

\endlocaldefs

\begin{document}

\begin{frontmatter}

\title{Topic-adjusted visibility metric for scientific articles}
\runtitle{Topic-adjusted visibility metric for scientific articles}




\begin{aug}
  \author{\fnms{Linda S. L.}  \snm{Tan}\thanksref{m1,m2}\ead[label=e1]{st2924@columbia.edu}},
  \author{\fnms{Aik Hui} \snm{Chan}\thanksref{m2}\ead[label=e2]{phycahp@nus.edu.sg}}
  \and
  \author{\fnms{Tian}  \snm{Zheng}\thanksref{t1,m1}
  \ead[label=e3]{tzheng@stat.columbia.edu}%
  \ead[label=u1,url]{http://www.foo.com}}

  \thankstext{t1}{Corresponding author: tzheng@stat.columbia.edu}
  \runauthor{Tan, Chan and Zheng}
  \affiliation{Columbia University\thanksmark{m1} and National University of Singapore\thanksmark{m2}}

  \address{Department of Statistics \\
  Columbia University \\
  Room 1005 SSW, MC 4690 \\
  1255 Amsterdam Avenue \\
  New York, NY 10027 \\
          \printead{e3} \\
           \printead{e1}}
           

  \address{Physics Department \\
  Blk S12 (Mezzanine Level) \\
  Faculty of Science \\
  National University of Singapore \\
  2 Science Drive 3 \\
  Singapore 117551\\ 
          \printead{e2}}
\end{aug}

\begin{abstract}
Measuring the impact of scientific articles is important for evaluating the research output of individual scientists, academic institutions and journals. While citations are raw data for constructing impact measures, there exist biases and potential issues if factors affecting citation patterns are not properly accounted for. In this work, we address the problem of field variation and introduce an article level metric useful for evaluating individual articles' visibility. This measure derives from joint probabilistic modeling of the content in the articles and the citations amongst them using latent Dirichlet allocation (LDA) and the mixed membership stochastic blockmodel (MMSB). Our proposed model provides a visibility metric for individual articles adjusted for field variation in citation rates, a structural understanding of citation behavior in different fields, and article recommendations which take into account article visibility and citation patterns. We develop an efficient algorithm for model fitting using variational methods. To scale up to large networks, we develop an online variant using stochastic gradient methods and case-control likelihood approximation. We apply our methods to the benchmark KDD Cup 2003 dataset with approximately 30,000 high energy physics papers. 
\end{abstract}



\end{frontmatter}

\section{Introduction}
Measuring the impact and influence of scientific articles is important for evaluating the work of individual scientists \citep{Hirsch2005, Abramo2011} and comparing journals \citep{Garfield2006, Moed2010}. For researchers, such information is one of the most considered factors for hiring, promotion, funding decisions, award consideration and professional recognition. For academic journals, it is an indicator of a journal's stature among its peers, which is valuable for various reasons, from being considered by prospective authors for paper submission to being sought after by readers who need authoritative opinions on a topic.

Due to the lack of a unified definition, the quality and importance of scientific articles are often judged based on the journal in which they are published \citep{Simons2008}. In particular, a journal's {\it impact factor} \citep[see][]{Garfield2006}, defined using a journal's average number of citations per article, is frequently used as an indicator of the ``quality" of its articles and a means of evaluating the research output of individuals and institutions \citep{Casadevall2014}. However, studies have shown that such usage can be misleading; the journal impact factor conceals differences in citation rates among articles, is research field dependent and does not measure the scientific quality of individual articles \citep{Seglen1997}. To improve the assessment of scientific research by academic institutions, funding agencies and other parties, the \textit{San Francisco declaration on research assessment}\footnote{Outcome of a gathering of scientists at the Annual meeting of the American Society for Cell Biology on December 2012} recommends (among other proposals) placing greater emphasis on the scientific content of an article rather than the journal impact factor \citep[see][]{Alberts2013}. An increasing number of publishers and organizations are also providing article level metrics \citep{Neylon2009, Fenner2014} to enable users to gauge the impact of articles based on their own merits. These new indicators include data on usage activity, bookmarks (e.g. CiteULike and Mendeley) and discussions/recommendations on the Social Web (e.g. Twitter, Facebook, Blogs) in addition to citations. 

Citations (and other reference counts alike) are raw data for constructing measures to evaluate the impact of scientific articles. The $h$-index \citep{Hirsch2005}, for instance, attempts to measure the impact of an author's published work using citations (a researcher who has published $h$ papers each having at least $h$ citations has index $h$). However, there exist biases and potential issues in using raw citations to compare the impact of scientific articles without accounting for other factors which may affect citation patterns. These factors include time from publication, journal profile, article type and social network of authors \citep{Bornmann2008}. A well-known and highly relevant factor is the variation in citation practices among different disciplines \citep{Garfield1979}. Articles in certain disciplines (e.g. Social Science and Mathematics) are typically much less cited than others (e.g. Molecular Biology and Immunology) and comparing articles using raw citation counts would be inappropriate. To address this issue, different procedures of normalizing the citation counts with respect to some reference standard have been proposed \citep[e.g.][]{Schubert1996, Vinkler2003, Radicchi2008}.  More recently, \cite{Crespo2013a} and \cite{Crespo2013b} consider a model where the number of citations received by an article depends on the subfield to which the article belongs and the scientific influence of the article in the subfield. Their model assumes that citation impact varies monotonically with scientific influence. 

In this work, we introduce an article level measure (of citation likelihood) that accounts for the variation in citation practices in different fields and is potentially useful for evaluating the impact of scientific articles. This measure, named as {\it topic-adjusted visibility metric}, derives from joint probabilistic modeling of the content (text) in the articles and the citations (links) amongst them. We consider a framework whereby the connectivity of an article in a citation network depends on (1) the citation probability of the research fields (topics) that it belongs to and (2) its \textit{visibility} to articles that are in a position to cite it. Our motivation is that while a citation is driven primarily by compatibility in research topics, the decision to cite an article over other equally relevant ones may be due to a complex mixture of attributes of the selected article which are unobserved/hard to quantify (e.g. research value and quality, profile of authors, journal readership) or difficult to model directly (e.g. time since publication, article type). Here we use the term \textit{visibility} to capture collectively attributes of the article apart from research topics that accounts for its connectivity. In our model, the topics are discovered using only the text and connectivity information. It does not take into account the discipline classification of the article by the journal. We also use the term ``field" to refer to a particular topic (area).

As citation networks are a type of relational data where content information is available on individual nodes, our proposed model combines two well-established models (for text and relational data respectively): latent Dirichlet allocation \citep[LDA,][]{Blei2003} and the mixed membership stochastic blockmodel \citep[MMSB,][]{Airoldi2008}. LDA is a generative probabilistic model which can uncover research topics from the text of scientific articles, while the MMSB can detect communities within the citation network and model inter- and intra-community citation probabilities. As the communities detected in citation networks often correlate well with research topics \citep{Chen2010}, these two models can be integrated by identifying the communities in MMSB with the topics in LDA \cite[Pairwise-Link-LDA,][]{Nallapati2008}. We further introduce a latent variable at the article level into the MMSB, which scales the probability of a citation due to compatibility in research topics and acts as a measure of the visibility of individual articles. The proposed model provides a structural understanding of the field variation in citation behavior and a measure of visibility for individual articles adjusted for citation probabilities within/between topics. 

Our model can also provide article recommendations which take into account individual articles' visibility and citation patterns across different topics. Consider a scenario where one is searching for papers on a computational technique applied in multiple topic areas by using keywords. A method which sorts relevant articles by citation counts may yield a list where papers in topics with higher citation rates are overrepresented at the top. We avoid this scenario as articles are recommended based on the citation probability within/across topics as well as the visibility metric which has adjusted for field variation in citation behavior. Hence, high impact articles in topics with low citation rates will not be overlooked. As the MMSB is able to capture both inter- and intra-topic citation probabilities, relevant articles which integrate multiple topics can also be identified. 

The proposed topic-adjusted visibility metric is novel and differs from approaches based on normalization of citation counts. While similar in motivation with \cite{Crespo2013a} and \cite{Crespo2013b}, our model is significantly different from theirs. First, they do not consider the text of articles and make use of an external system provided by Thomson Reuters for classification (which may be limited in range) while we identify research topics in the articles using LDA and MMSB jointly. Moreover, our model does not assume that citation counts vary monotonically with visibility within each topic and is fully generative for text and citations. Besides citations, other reference data (e.g. usage activity) which are field dependent can also benefit from our proposed framework.

For model fitting, we adopt a Bayesian approach and develop efficient variational methods \citep{Jordan1999} for fast approximate posterior inference. As real-world citation networks are often massive and the computational cost of analyzing every pairwise interaction in the MMSB scales as the square of the number of nodes, we develop an online variant of our variational algorithm by subsampling the full network using case-control likelihood approximation techniques \citep{Raftery2012} and stochastic variational inference \citep{Hoffman2013}. Previously, stochastic variational inference has been employed successfully for LDA \citep{Hoffman2010} and the hierachical Dirichlet process \citep{Wang2011a}. At each iteration, it subsamples the data and optimizes the variational objective using stochastic approximation methods \citep{Robbins1951}, thus reducing both computational and storage costs. Recently, \cite{Gopalan2013} extend stochastic variational inference to massive networks for detecting overlapping communities by using a variant of the MMSB. They sample node pairs using ``informative set sampling", where the sets of pairs are defined using  network topology information. A related idea is the stratified sampling scheme for MCMC estimation of latent space models \citep{Raftery2012}, where stratums are defined by shortest path lengths. Adapting case-control designs in epidemiology, they approximate the log-likelihood function by sampling all links for each node and only a small proportion of nonlinks from each stratum. This approach is feasible as large networks are often sparse. It assumes that ``closer" nodes contain more information and are more relevant in estimating each other's latent position. Motivated by these methods, we propose a novel strategy for sampling node pairs that is suitably adapted to our model requirements. 

We apply our methods to the Cora dataset with 2410 scientific publications in computer science research and the benchmark KDD Cup 2003 dataset with approximately 30,000 high energy physics papers. We also evaluate the performance of our model using a simulation study. A particularity of citation networks is that articles join the network over time, and published articles cannot cite articles appearing at a later date. Hence the absence of such links cannot be construed as true ``zeros" and should be omitted from the likelihood. We show that taking into account publication times (when available) can significantly improve performance of our model.

The rest of the paper is organized as follows. Section \ref{LMV} lays out the details of our model and reviews closely associated models. Section \ref{posterior inference} introduces a variational algorithm for obtaining approximate posterior inference and Section \ref{stochastic optimization} describes how the algorithm can be scaled up to large networks using stochastic optimization methods. Section \ref{Comparison} discusses comparisons with alternative approaches and Section \ref{citation prediction} predictions and article recommendations based on our proposed model. Section \ref{examples} presents application results using simulations and real data. We conclude with discussion in Section \ref{conclusion}.

\section{Model description} \label{LMV}
Our proposed model combines LDA with the MMSB, and introduces a latent variable for each article which acts as a measure of its visibility adjusted for topic-level variation in activity level. Before describing our model, we review LDA, MMSB and other associated models.

\subsection{Review of LDA, MMSB and Pairwise-Link-LDA} \label{review PLLDA}
LDA is a generative probabilistic model for text copora which can be used for tasks such as detecting themes, summarization and classification. It assumes that word order can be disregarded (``bag-of-words" model) and that each document in the corpus exhibits $K$ topics  with varying proportions. Let the number of documents in the corpus be $D$ and the size of the vocabulary be $\mathcal{V}$. Each topic $\beta_k$ is a $\mathcal{V} \times 1$ vector with a Dirichlet($\eta$) prior, representing a probability distribution over the vocabulary. For each document $d$, the topic proportions $\theta_d$ is a $K \times 1$ vector with a Dirichlet($\alpha$) prior, representing the probability of each topic occurring in the document. Let the number of words in document $d$ be $N_d$. The $n$th word in document $d$, $w_{dn}$, is generated by drawing a topic assignment $z_{dn}$ from $\text{Multinomial} (\theta_d)$ and the word from $\text{Multinomial} \left( \beta_{z_{dn}} \right)$. Both $z_{dn}$ and $w_{dn}$ are indicator vectors with a single one. If the $k$th element of $z_{dn}$ is one, $w_{dn}$ is drawn from the topic $\beta_{z_{dn}}$, which refers to $\beta_k$ (slight abuse of notation).

On the other hand, MMSB is a mixed membership model for relational data that can detect communities within a network. Suppose the relational data is represented by a directed graph. For each node pair $(d,d')$ where $d \neq d'$, we define the binary variable $y_{dd'}$ to be 1 if  there is a directed edge from $d$ to $d'$ and 0 otherwise. The MMSB assumes that there are $K$ latent communities (groups) and each node belongs to the $K$ groups with varying degrees of affiliation. Specifically, each node $d$ is associated with a $K \times 1$ membership vector, $\theta_d$, drawn from a Dirichlet($\alpha$) prior, representing the probability of the node belonging to each of the $K$ groups. Each node may assume different membership when interacting with different nodes. The blockmodel $B$ is a $K \times K$ matrix where $B_{ij}$ represents the probability of a directed link from a node in group $i$ to a node in group $j$. For each $(d,d')$, membership indicator vectors for the {\em sender} ($s_{dd'}$) and {\em receiver} ($r_{dd'}$) are first drawn from $\text{Multinomial} (\theta_d)$ and $\text{Multinomial} (\theta_{d'})$ respectively. If the $i$th and $j$th elements of $s_{dd'}$ and $r_{dd'}$ are ones respectively, the value of the interaction $y_{dd'}$ is sampled from Bernoulli$(B_{s_{dd'} r_{dd'}})$ where $B_{s_{dd'} r_{dd'}}$ refers to $B_{ij}$. The generating process of LDA and MMSB are shown in Figure \ref{LDA_MMSB}.

\begin{figure}
\begin{footnotesize}
\begin{minipage}{0.49\textwidth}
\centering
\parbox{6.2cm}{
\hrule \vspace*{0.7mm}
\textbf{LDA} (for modeling text)
\vspace{0.7mm} \hrule
\begin{enumerate}[noitemsep,leftmargin=1em,topsep=3pt, labelsep=3pt]
\item Draw topic $\beta_k \sim \text{Dirichlet} (\eta)$ for \\$k=1, \dots, K$.
\item For each document $d=1, \dots, D$,
\begin{enumerate}[noitemsep,topsep=1pt,leftmargin=0.8em]
\item draw topic proportion $\theta_d \negmedspace \sim  \negmedspace \text{Dirichlet} (\alpha) $.
\item For each position $n=1,\dots,N_d$, draw
\begin{itemize}[noitemsep,topsep=1pt,leftmargin=0.7em]
\item topic assignment \\
$z_{dn} \sim \text{Multinomial} (\theta_d) $.
\item word $w_{dn} \sim \text{Multinomial} \left( \beta_{z_{dn}} \right) $.
\end{itemize}
\end{enumerate}
\end{enumerate}
\hrule }
\end{minipage}
\begin{minipage}{0.49\textwidth}
\centering
\parbox{6.35cm}{
\hrule \vspace*{0.7mm}
\textbf{MMSB} (for modeling links)
\vspace{0.7mm} \hrule
\begin{enumerate}[noitemsep,leftmargin=1em,topsep=3pt, labelsep=3pt]
\item For each node $d=1, \dots, D$, draw mixed membership vector $\theta_d \sim \text{Dirichlet} (\alpha) $.
\item For each node pair $(d, d')$, draw
\begin{itemize}[noitemsep,topsep=1pt,leftmargin=0.7em]
\item membership indicator for sender \\
$s_{dd'} \sim \text{Multinomial} (\theta_d)$.
\item membership indicator for receiver \\
$r_{dd'} \sim \text{Multinomial} (\theta_{d'}) $.
\item interaction $y_{dd'}\sim \text{Bernoulli} (B_{s_{dd'} r_{dd'}} )$.
\end{itemize}
\end{enumerate}
\hrule }
\end{minipage}
\caption{Generating process for LDA (left) and MMSB (right). These two models can be combined by identifying the communities in MMSB with the topics in LDA, that is, the topic proportions with the mixed membership vectors (both denoted by $\theta_d$).}
\label{LDA_MMSB} 
\end{footnotesize}
\end{figure}

Citation networks are a type of relational data where the nodes are the documents/articles and the directed links are the citations between them (there is a directed link from $d$ to $d'$ if $d$ cites $d'$). Pairwise-Link-LDA \citep{Nallapati2008} combines MMSB with LDA to jointly model text in articles and the citations between them by identifying the topics in LDA with the communities in MMSB. As links between documents indicate a certain level of similarity in topics, it is believed that network information, when suitably incorporated, would improve topic modeling \citep{Kleinberg1999, Ho2012b}. 

\subsection{Proposed model: LMV} \label{S proposed model}

\begin{figure}
\begin{minipage}{0.55\textwidth}
\centering
\parbox{7cm}{
\footnotesize
\hrule \vspace*{0.7mm}
\textbf{Generative process of LMV}
\vspace{0.5mm} \hrule
\begin{enumerate}[noitemsep,leftmargin=1.5em]
\item Draw topic $\beta_k \sim \text{Dirichlet} (\eta) \;\text{for} \; k=1, \dots, K$.
\item For each document $d=1, \dots, D$,
\begin{itemize}[noitemsep,topsep=1pt,leftmargin=1em]
\item draw visibility $\tau_{d} \sim \text{Beta}(g_0, h_0)$.
\item draw topic proportion $\theta_d \sim \text{Dirichlet} (\alpha) $.
\item For each position $n=1,\dots,N_d$, draw
\begin{itemize}[noitemsep,topsep=1pt,leftmargin=1em]
\item topic assignment $z_{dn} \sim \text{Multinomial} (\theta_d) $.
\item word $w_{dn} \sim \text{Multinomial} \left( \beta_{z_{dn}} \right) $.
\end{itemize}
\end{itemize}
\item For $i, j \in \{ 1, \dots, K\}$, draw $B_{ij} \sim \text{Beta}(a_0, b_0)$.
\item For each document pair $(d, d')$ where  $d \neq d'$,
\begin{itemize}[noitemsep,topsep=1pt,leftmargin=1em]
\item draw topic indicator for \\
sender: $s_{dd'}  \sim \text{Multinomial} (\theta_d)$. \\
receiver: $r_{dd'}  \sim \text{Multinomial} (\theta_{d'}) $.
\item draw $y_{dd'} \sim \text{Bernoulli} (\tau_{d'} B_{s_{dd'} r_{dd'}} )$.
\end{itemize}
\end{enumerate}
\hrule }
\end{minipage}
\hspace{4mm}
\begin{minipage}{0.37\textwidth}
\centering
\includegraphics[width=\textwidth]{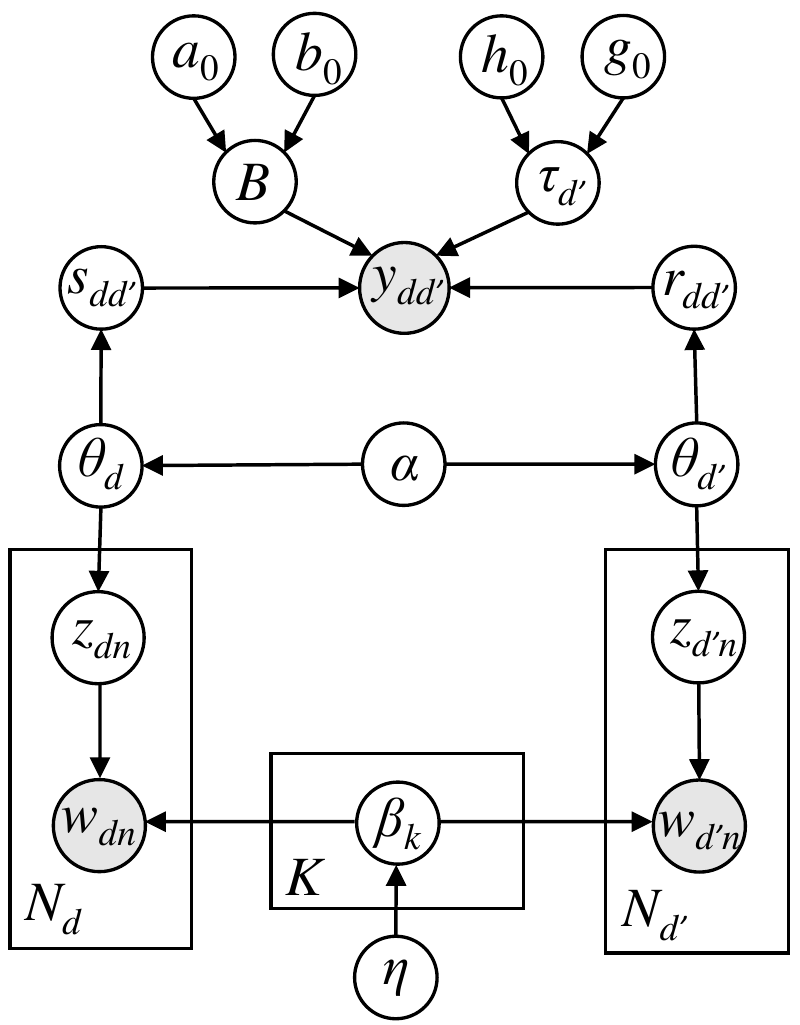}
\end{minipage} \hspace{4mm}
\caption{Left: Outline of LMV. Right: Graphical representation of a two-document segment. The complete model contains $y_{dd'}$ for every document pair. Circles denote variables and observed variables are shaded. The plates contain variables to be replicated and the number of times is indicated in the lower left corner.}
\label{LMV figure} 
\end{figure}

In Pairwise-Link-LDA, any two documents with the same topic proportions have equal probability of being cited. This assumption is easily violated in real-world citation networks as factors other than research topics affect the citation probability. For instance, a well-cited document's higher chances of being cited may be due to its quality, novelty and the authors' social networks. Our proposed model aims to capture collectively via the \textit{visibility} measure, attributes of the cited document that explains this variation in citation probability given compatibility in topics.

Our proposed model is presented in Figure \ref{LMV figure}. The text is generated as in LDA. For the generation of links, we introduce a latent variable $\tau_{d'}$ for each document $d'$, drawn from a $\text{Beta}(g_0, h_0)$ prior, which modifies the citation probability by scaling the blockmodel. Given that the $i$th element of $s_{dd'}$ and $j$th element of $r_{dd'}$ are ones, the probability of a document $d$ from the $i$th topic citing a document $d'$ from the $j$th topic becomes $\tau_{d'} B_{ij}$, which is dependent on the characteristic $\tau_{d'}$ of the document $d'$ receiving the citation. 

We define $\tau_{d'}$ as the \textit{topic-adjusted visibility} of document $d'$, which is a scaling factor that adjusts the universal probability of being cited based on citation probabilities within/between topics. It is a characteristic of $d'$ that accounts for the variation in citation probability among documents with equal topic proportions. This variation might be due to attributes of $d'$ which are unobserved, hard to quantify, or not directly modeled. The topic-adjusted visibility is potentially useful as a descriptive article level measure of citation likelihood that adjusts for the differences in citation practices in different topic areas. We use the acronym LMV for our model, taking the first letter from LDA, MMSB and visibility. 

Our model reduces to Pairwise Link-LDA \citep{Nallapati2008} when $\tau_{d'}$ is identically one. One computational issue associated with LDA and MMSB is the multi-modality of these models, whereby multiple model configurations give equivalent fits to the data \citep[see e.g.][]{Ho2012a}. Combining content and connectivity information in a citation network may alleviate the multi-modality issue of MMSB. On the other hand, the communities detected in the citation network regularize the topic modeling on the content information.

We assume that the hyperparameters $a_0$, $b_0$, $g_0$, $h_0$, $\eta$ and $\alpha$ are known. The latent variables $s_{dd'}$ and $r_{dd'}$ give rise to a more tractable joint distribution, which leads to significant simplification of the variational optimization procedure to be introduced in Section \ref{posterior inference}.

\subsection{Review of associated methods} \label{S review methods}
The relational topic model \citep[RTM,][]{Chang2010} also uses LDA as a basis for modeling document networks. RTM does not consider every pairwise interaction, however, and models only observed links. It uses a symmetric probability function with a diagonal weight matrix, which allows only within topic interactions. To improve the RTM, extensions have been proposed. For instance, \cite{Chen2013} define generalized RTM with a full weight matrix and perform regularized Bayesian inference where a log-logistic loss is minimized. A regularization parameter is used to control influence from link structures. \cite{Zhang2013} propose sparse RTM where normal and Laplace priors are placed respectively on the topic and word representations, and word counts are Poisson distributed. Different regularization parameters are used for links and non-links in the minimization of a log-loss. 

Some other extensions of LDA to document networks include the TopicBlock \citep{Ho2012b}, which uses text and links to induce a hierarchical taxonomy, and block LDA \citep{Balasubramanyan2013}, which considers documents annotated with entities and models only realized links between entities using a stochastic blockmodel. \cite{Zhu2013} propose a Poisson mixed-topic link model that combines LDA with a variant of the MMSB, called the Ball-Karrer-Newman model, where the number of links between two documents is Poisson distributed instead of Bernoulli. \cite{Neiswanger2014} present a latent random offset model that augments the topic proportions of the cited document with a vector to capture contents of citing documents in link predictions. These models do not address the issue that documents with similar topics may have different connectivity due to unobserved factors of impact.

To improve the understanding of large text corpora, models that incorporate additional information about the corpus or document metadata into text analysis have also been introduced recently. These models investigate the relationship between text data and observed variables that may affect the text composition (e.g. authors, date, political affiliation and quality ratings). To overcome the difficulty of incorporating high-dimensional text data into statistical analyses for predicting sentiment variables, multinomial inverse regression \citep{Taddy2013} uses the inverse conditional distribution of text given sentiment to obtain low-dimensional document representations that preserve sentiment information. The inverse regression topic model \citep{Rabinovich2014} extends multinomial inverse regression to the mixed-membership (multiple topics) setting, while distributed multinomial regression \citep{Taddy2015} tackles high-dimensional sentiment variables by modeling document-word counts as independent Poisson distributed variables. \cite{Roberts2013} propose the structural topic model, which uses generalized linear models as priors to incorporate document-level covariates. In this paper, we adopt a different approach to the exploration of large text copora by modeling text and links between documents jointly. Further incorporating document metadata into our proposed model will be an interesting direction for future research.

\section{Posterior Inference} \label{posterior inference}
As the true posterior of our model is not available in closed form, we develop an efficient variational algorithm for posterior approximation. Let $\Theta$ denote the set of unknown variables in the LMV. In variational methods, the true posterior is approximated by more tractable distributions which are optimized to be close to the true posterior in terms of Kullback-Leibler divergence. Here we consider a fully-factorized family,
\begin{multline*}
q(\Theta) = \prod_d \Big\{ q_D(\theta_d|\gamma_d)  \prod_n q_M(z_{dn}|\phi_{dn}) \prod_{d' \neq d} [q_M(s_{dd'}|\kappa_{dd'})q_M(r_{dd'}| \nu_{dd'}) ] \Big\} \\
\times \prod_k q_D(\beta_k|\lambda_k) \prod_{i, j} q_B(B_{ij}|a_{ij},b_{ij})  \prod_{d'} q_B(\tau_{d'}|g_{d'},h_{d'}),
\end{multline*}
where $q_D$, $q_M$, $q_B$ denote the Dirichlet, multinomial and beta distributions respectively and $\{ \lambda, \gamma, \phi, \kappa, \nu, a, b, g, h \}$ are variational parameters to be optimized. Discussion on assumptions made in the variational approximation and their implications can be found in the \ref{suppA}.

From Jensen's inequality, minimizing the Kullback-Leibler divergence between $q(\Theta)$ and the true posterior is equivalent to maximizing a lower bound $\mathcal{L}$ on the log marginal likelihood, where $\mathcal{L}$ is given by
\begin{multline}\label{lower bound}
\sum_{(d, d')} \text{E}_q  [\log p(y_{dd'}| \tau_{d'}, B, s_{dd'}, r_{dd'})+ \log p(s_{dd'}| \theta_d) + \log p(r_{dd'}| \theta_{d'})] \\
\hspace*{7mm}+  \sum_{d,n}  \text{E}_q [\log p(z_{dn}| \theta_d) +  \log p(w_{dn} |z_{dn}, \beta)]  + \sum_{i, j} \text{E}_q [\log p(B_{ij}| a_0, b_0)] \\
+ \sum_{d} \text{E}_q [\log p(\tau_{d}| g_0, h_0)  + \log p(\theta_d|\alpha)]  + \sum_k \text{E}_q [\log p(\beta_k|\eta)] + H(q). 
\end{multline}
In \eqref{lower bound}, $\text{E}_q$ denotes expectation with respect to $q(\Theta)$ and $H(q)$ denotes the entropy of $q$. All terms in $\mathcal{L}$ can be evaluated analytically except $\text{E}_q\{ \log (1-\tau_{d'} B_{ij}) \}$. We expand this expectation using a first-order approximation about the mean \citep{Braun2010} so that 
\begin{equation} \label{first-order approx}
\begin{aligned}
\text{E}_q\{ \log (1-\tau_{d'} B_{ij}) \}  & \approx  \log (1 - \text{E}_q(\tau_{d'}) \text{E}_q(B_{ij})) \\
& = \log \Big( 1 - \frac{g_{d'}}{g_{d'} + h_{d'}} \frac{a_{ij}}{a_{ij} + b_{ij}} \Big).
\end{aligned}
\end{equation}
The approximate lower bound obtained using \eqref{first-order approx} is denoted by $\mathcal{L}^*$. Discussion on the first-order approximation and the expression for $\mathcal{L}^* $ can be found in the \ref{suppA}. 

We optimize $\mathcal{L}^*$ with respect to the variational parameters via coordinate ascent (see Algorithm 1). For $\{ \lambda, \gamma, \phi, \kappa, \nu \}$, closed form updates can be derived by differentiating $\mathcal{L}^*$ with respect to each parameter and setting the gradient to zero. For $\{ a, b, g , h \}$, the likelihood is nonconjugate with respect to the prior and we use nonconjugate variational message passing \citep{Knowles2011}. This is a fixed point iteration method for optimizing the natural parameters of variational posteriors in exponential families. The advantages of this approach is that it yields closed form updates and extends to stochastic variational inference naturally. However, $\mathcal{L}^*$ is not guaranteed to increase at each step and updates for $\{ a, b, g , h \}$ may be negative at times. To resolve these issues, we use the fact that nonconjugate variational message passing is a natural gradient ascent method with step size 1 and smaller step sizes may also be taken. In Algorithm 1, we start with step size 1 and reduce the step size where necessary to ensure updates of $\{ a, b, g , h \}$ are positive. If $\mathcal{L}^*$ increases, these updates are accepted. Otherwise, we revert to the former values. Updates for $\{ a, b, g , h \}$ are derived in the \ref{suppA}. As updates of $\{a, b\}$ and $\{g, h\}$ are highly interdependent, we introduce a nested loop for cycling these updates in step 5 of Algorithm 1.
\begin{algorithm}
\small
\vspace{1mm}
\begin{flushleft}
Initialize $\lambda$, $\gamma$, $\phi$, $\kappa$, $\nu$, $a$, $b$, $g$ and $h$. Cycle the following updates until convergence is reached.
\end{flushleft}
\begin{enumerate}[noitemsep,leftmargin=1.3em,topsep=1pt,parsep=1pt,partopsep=1pt]
\item For each document pair $(d,d')$, cycle the following updates until $\kappa_{dd'}$ and $\nu_{dd'}$ converge.
\begin{equation*}
\begin{aligned}
\kappa_{dd'i} & \propto \exp \left\{ \psi(\gamma_{di})-\psi\left(\sum\nolimits_i \gamma_{di}\right) +  \sum\nolimits_j \nu_{dd'j} \varsigma_{dd'}(i,j)  \right\}  \; \text{for} \; i=1, \dots, K, \\
\nu_{dd'j} &\propto \exp \left\{ \psi(\gamma_{d'j})-\psi\left(\sum\nolimits_j \gamma_{d'j}\right) +  \sum\nolimits_i \kappa_{dd'i} \varsigma_{dd'}(i,j)  \right\}  \; \text{for} \; j=1, \dots, K,
\end{aligned}
\end{equation*}
\vspace{-3mm}
\begin{equation*}
\text{where} \; \varsigma_{dd'}(i,j) =  \begin{cases} \psi(a_{ij}) - \psi(a_{ij} + b_{ij}) + \psi(g_{d'}) - \psi(g_{d'} + h_{d'}) & \text{if} \; y_{dd'} = 1, \\
\log \left(1- \frac{g_{d'}}{g_{d'}+h_{d'}} \frac{a_{ij}}{a_{ij} +b_{ij}} \right) & \text{if} \; y_{dd'} = 0. \end{cases}
\end{equation*}
\item For $d=1, \dots, D$, $n=1, \dots, N_d$, $k=1, \dots, K$, 
\begin{equation*}
\phi_{dnk} \propto \exp\left\{  \psi(\gamma_{dk})-\psi\left(\sum\nolimits_k \gamma_{dk}\right)+ \sum\nolimits_v w_{dnv}  \left[\psi(\lambda_{kv})-\psi\left(\sum\nolimits_v \lambda_{kv}\right)\right]  \right\}.
\end{equation*}
\item For $d=1, \dots, D$, $\gamma_d = \alpha +  \sum_n \phi_{dn} +   \sum_{d' \neq d} ( \kappa_{dd'} + \nu_{d'd} ) $.
\item For $k=1, \dots, K$, $v=1, \dots, V$, $\lambda_{kv} = \eta_v  + \sum_d \sum_n  w_{dnv} \phi_{dnk}$.
\item Cycle updates in (a) and (b) until convergence is reached.
\begin{enumerate}[noitemsep,leftmargin=1.3em,topsep=1pt]
\item For $i=1, \dots, K$, $j=1, \dots, K$, $\begin{bmatrix} a_{ij} \\ b_{ij} \end{bmatrix} \leftarrow (1- s_t) \begin{bmatrix} a_{ij} \\ b_{ij} \end{bmatrix}  + s_t  \begin{bmatrix} \hat{a}_{ij} \\ \hat{b}_{ij} \end{bmatrix}$, where
\begin{multline*}
\begin{bmatrix} \hat{a}_{ij} \\ \hat{b}_{ij} \end{bmatrix} = \begin{bmatrix} a_0 +  \sum_{ (d,d'):\; y_{dd'}=1} \kappa_{dd'i}\nu_{dd'j}  \\ b_0 \end{bmatrix}   + \frac{1}{|I_{a_{ij}, b_{ij}}|(a_{ij} +b_{ij})^2} \\  \times 
\begin{bmatrix} (a_{ij} + b_{ij}) \psi'(a_{ij} + b_{ij}) - b_{ij} \psi'(b_{ij}) \\ a_{ij} \psi'(a_{ij}) -(a_{ij} + b_{ij}) \psi'(a_{ij} + b_{ij}) \end{bmatrix}  \sum_{ (d,d'):\; y_{dd'}=0} \frac{\kappa_{dd'i}\nu_{dd'j}\frac{g_{d'}}{g_{d'}+h_{d'}} }{1- \frac{g_{d'}}{g_{d'}+h_{d'}} \frac{a_{ij}}{a_{ij} +b_{ij}} }.
\end{multline*}
Start with $s_t=1$. If any $a_{ij} \leq 0$ or $b_{ij} \leq 0$, reduce $s_t$ (say by half each time) until all $a_{ij} > 0$ and $b_{ij} > 0$. Accept update only if $\mathcal{L}^*$ increases. 
\item For $d'=1, \dots, D$, $\begin{bmatrix} g_{d'} \\ h_{d'} \end{bmatrix} \leftarrow (1- s_t) \begin{bmatrix} g_{d'} \\ h_{d'} \end{bmatrix}  + s_t  \begin{bmatrix} \hat{g}_{d'} \\ \hat{h}_{d'} \end{bmatrix} $, where
 \begin{multline*}
\begin{bmatrix}\hat{g}_{d'} \\ \hat{h}_{d'} \end{bmatrix} = \begin{bmatrix} g_0 + \sum_d y_{dd'} \\ h_0 \end{bmatrix}   
+ \frac{1}{|I_{g_{d'}, h_{d'}}|(g_{d'} +h_{d'})^2}  \\ \times 
\begin{bmatrix} (g_{d'} + h_{d'}) \psi'(g_{d'} + h_{d'}) - h_{d'} \psi'(h_{d'}) \\ g_{d'} \psi'(g_{d'}) -(g_{d'} + h_{d'}) \psi'(g_{d'} + h_{d'}) \end{bmatrix}  \sum_{i, j}       \frac{  \sum_{ d:\; y_{dd'}=0}   \kappa_{dd'i}\nu_{dd'j}  \frac{a_{ij}}{a_{ij} +b_{ij}}  }{1- \frac{g_{d'}}{g_{d'}+h_{d'}} \frac{a_{ij}}{a_{ij} +b_{ij}} }.
\end{multline*}
Start with $s_t=1$. If any ${g}_{d'} \leq 0$ or ${h}_{d'} \leq 0$, reduce $s_t$ (say by half each time) until all ${g}_{d'} > 0$ and ${h}_{d'} > 0$. Accept update only if $\mathcal{L}^*$ increases. 
\end{enumerate}
\end{enumerate}
\begin{flushleft}
Note: $|I_{a,b}|$ denotes determinant of the Fisher information matrix of $\text{Beta}(a,b)$. See \ref{suppA}.
\end{flushleft}
\caption{Coordinate ascent procedure for the LMV}
\label{Algorithm 1}
\end{algorithm}

\section{Stochastic optimization of variational objective} \label{stochastic optimization}
We develop an online variant of Algorithm 1 that scales well to large networks using stochastic variational inference \citep{Hoffman2013}. In this approach, variational parameters are classified as {\it local} (specific to each node) or {\it global} (common across all nodes) parameters. At each iteration, a minibatch of nodes are randomly sampled from the whole dataset and local parameters corresponding to these nodes are optimized. Global parameters are then updated based on optimized local parameters using stochastic gradient ascent \citep{Robbins1951}. The algorithm converges to a local maximum of the variational objective provided the step sizes and the objective function satisfy certain regularity conditions \citep[see][]{Spall2003}.

Currently, Algorithm 1 has to update the variational parameters $\kappa_{dd'}$ and $\nu_{dd'}$ for each document pair $(d,d')$ at every iteration. This computational cost scales as $\mathcal{O}(D^2)$ and makes our model infeasible for large networks. To apply stochastic variational inference, we regard $\kappa_{dd'}$ and $\nu_{dd'}$ as {\it local} parameters and perform these updates only for a random subset of all document pairs at each iteration. Remaining variational parameters are regarded as {\it global} parameters and are updated using stochastic gradient ascent.

\subsection{Proposed sampling strategy} \label{sampling_strategy}
We devise a novel scheme for sampling document pairs. While simple random sampling is a possibility, it does not utilize information provided by the links. \cite{Raftery2012} propose a stratified sampling scheme where stratums are defined by shortest path lengths. Assuming ``closer" nodes contain more information and that large networks are often sparse, they sample all links for each node and only a small proportion of nonlinks from each stratum. \cite{Gopalan2013} consider ``informative set sampling", where ``informative set" for a node consists of all links and nonlinks of path length 2. Remaining nonlinks are partitioned into ``noninformative sets". At each iteration, either an ``informative" set is chosen with high probability or one of the ``noninformative sets" is chosen with low probability. These schemes are not directly applicable to our model. To update $\gamma_d$ (Algorithm 1, step 2), unbiased estimates of $\sum_{d'\neq d}  \kappa_{dd'}$ and $\sum_{d' \neq d} \nu_{d'd}$ are required. That is, samples has to be drawn from cases where $d$ is the citing document as well as cases where $d$ is the cited document. As \cite{Gopalan2013} treat links as undirected while \cite{Raftery2012} do not subsample documents, they do not face these restrictions. 

We propose the following sampling scheme. Consider the adjacency matrix $y$ of ones and zeros, where the rows and columns denote the citing and cited documents respectively (diagonal is undefined). We associate each document pair $(d,d')$ with an inclusion probability $\pi_{dd'}$, where $\pi_{dd'}$ is a decreasing function of the shortest path length from $d$ to $d'$. While other definitions are plausible, we define for simplicity,
\begin{equation} \label{inclusion prob}
\pi_{dd'} = \begin{cases} 1/l_{dd'} & \text{$l_{dd'} > n_0$}, \\
1/{n_0} & \text{otherwise}, \end{cases}
\end{equation}
where $l_{dd'}$ denotes the shortest path length from $d$ to $d'$ and $n_0$ is a positive integer. When $y_{dd'}=1$, $\pi_{dd'}=1$. Hence all links are included and more ``informative" nonlinks (in the sense of shorter path length) have a higher probability of being included. In the examples, we set $n_0=100$. This implies that document pairs with a shortest path length of 100 or more have a probability of 0.01 of being included. It is possible to experiment with other values depending on the application and computational constrains. A smaller $n_0$ implies a larger sample of document pairs. At each iteration, we
\begin{enumerate}[noitemsep,leftmargin=1.5em]
\item select a random sample $\mathcal{S}$ of $|\mathcal{S}|$ documents from the whole dataset.
\item perform a Bernoulli trial with success probability $\pi_{dd'}$ for each $(d,d')$ where $d \in \mathcal{S}$ or $d' \in \mathcal{S}$. 
\item select document pairs with successful trials (denote this set as $\mathcal{P}$). 
\end{enumerate}
The Bernoulli trial is performed only once for each document pair even if both $d$ and $d'$ are in $\mathcal{S}$.The sampling strategy is illustrated in the \ref{suppA}.

\subsection{Stochastic variational algorithm} \label{stochastic_gradient_updates}
\begin{algorithm}
\vspace{1mm}
\small
\begin{flushleft}
Initialize $\gamma$, $\lambda$, $\kappa$, $\phi$, $\nu$, $a$, $b$, $g$, $h$. At each iteration, 
\end{flushleft}
\begin{enumerate}[noitemsep,leftmargin=1.3em,topsep=1pt,parsep=1pt,partopsep=1pt]
\item Obtain a random sample $\mathcal{S}$ of $|\mathcal{S}|$ documents from the corpus.
\item For each document pair $(d,d')$, where $d \in \mathcal{S}$ or $d' \in \mathcal{S}$, perform a Bernoulli trial with success probability $\pi_{dd'}$. Let $\mathcal{P}$ denote the set of document pairs with successful trials. Let $\mathcal{P}_{d \Cdot} = \{(l,l') \in \mathcal{P} | l=d\}$ and $\mathcal{P}_{\Cdot d} = \{(l,l') \in \mathcal{P} | l'=d\}$.
\item Update $\kappa_{dd'}$ and $\nu_{dd'}$ iteratively as in Algorithm 1 for each $(d,d') \in \mathcal{P}$ until convergence.
\item Update $\phi_{dnk}$ for $d \in \mathcal{S}$, $n=1, \dots, N_d$ and $k=1, \dots, K$ as in Algorithm 1.
\item For $d \in \mathcal{S}$, $\gamma_d \leftarrow (1-s_t) \gamma_d + s_t \; \hat{\gamma}_d$ where
\begin{equation*}
\hat{\gamma}_d = \alpha +  \sum_n \phi_{dn} +  \sum_{(l,l')\in \mathcal{P}_{d \Cdot}} \frac{\kappa_{ll'}}{\pi_{ll'}} + \sum_{(l,l')\in \mathcal{P}_{\Cdot d}} \frac{\nu_{ll'}}{\pi_{ll'}}.
\end{equation*}
\item For $k=1, \dots, K$ and $v=1, \dots, V$,  $\lambda_{kv} \leftarrow (1-s_t) \lambda_{kv} + s_t  \hat{\lambda}_{kv}$, where
\begin{equation*}
\hat{\lambda}_{kv} = \eta_v  + \frac{D}{|\mathcal{S}|}\sum_{d \in \mathcal{S}} \sum_n w_{dnv} \phi_{dnk}.
\end{equation*}
\item For $i=1, \dots, K$ and $j=1, \dots, K$, $\begin{bmatrix} a_{ij} \\ b_{ij} \end{bmatrix} \leftarrow (1- s_t) \begin{bmatrix} a_{ij} \\ b_{ij} \end{bmatrix}  + s_t  \begin{bmatrix} \hat{a}_{ij} \\ \hat{b}_{ij} \end{bmatrix}$, where
\begin{multline*}
\begin{bmatrix} \hat{a}_{ij} \\ \hat{b}_{ij} \end{bmatrix} = \begin{bmatrix} a_0 +  \frac{D}{|\mathcal{S}|} \sum_{d' \in \mathcal{S}} \sum_{ (l,l') \in \mathcal{P}_{ \Cdot d'}: y_{ll'} =1} \kappa_{ll'i}\nu_{ll'j}  \\ b_0 \end{bmatrix}   
+ \frac{1}{|I_{a_{ij}, b_{ij}}|(a_{ij} +b_{ij})^2}  \\ \times 
\begin{bmatrix} (a_{ij} + b_{ij}) \psi'(a_{ij} + b_{ij}) - b_{ij} \psi'(b_{ij}) \\ a_{ij} \psi'(a_{ij}) -(a_{ij} + b_{ij}) \psi'(a_{ij} + b_{ij}) \end{bmatrix} \frac{D}{|\mathcal{S}|} \sum_{d' \in S} \frac{ \sum_{(l,l') \in\mathcal{P}_{\Cdot d'}: y_{ll'}=0} \frac{ \kappa_{ll'i}\nu_{ll'j}}{\pi_{ll'}} \frac{g_{d'}}{g_{d'}+h_{d'}} }{1- \frac{g_{d'}}{g_{d'}+h_{d'}} \frac{a_{ij}}{a_{ij} +b_{ij}} }.
\end{multline*}
If any $a_{ij} \leq 0$ or $b_{ij} \leq 0$, reduce $s_t$ for this update (say by half each time).
\item For $d' \in \mathcal{S}$, $\begin{bmatrix} g_{d'} \\ h_{d'} \end{bmatrix} \leftarrow (1- s_t) \begin{bmatrix} g_{d'} \\ h_{d'} \end{bmatrix}  + s_t  \begin{bmatrix} \hat{g}_{d'} \\ \hat{h}_{d'} \end{bmatrix} $, where 
\begin{multline*}
\begin{bmatrix} \hat{g}_{d'} \\ \hat{h}_{d'} \end{bmatrix} = \begin{bmatrix} g_0 +  \sum_d y_{dd'}  \\ h_0 \end{bmatrix}   
+ \frac{1}{|I_{g_{d'}, h_{d'}}|(g_{d'} +h_{d'})^2}  \\ \times 
\begin{bmatrix} (g_{d'} + h_{d'}) \psi'(g_{d'} + h_{d'}) - h_{d'} \psi'(h_{d'}) \\ g_{d'} \psi'(g_{d'}) -(g_{d'} + h_{d'}) \psi'(g_{d'} + h_{d'}) \end{bmatrix}  \sum_{i, j}       \frac{  \sum_{(l,l') \in \mathcal{P}_{\Cdot d'}: y_{ll'}=0}  \frac{ \kappa_{ll'i}\nu_{ll'j}}{\pi_{ll'}}  \frac{a_{ij}}{a_{ij} +b_{ij}}  }{1- \frac{g_{d'}}{g_{d'}+h_{d'}} \frac{a_{ij}}{a_{ij} +b_{ij}} }.
\end{multline*}
If any $g_{d'} \leq 0$ or $h_{d'} \leq 0$, reduce $s_t$ for this update (say by half each time).
\end{enumerate}
\caption{Stochastic variational procedure for the LMV}
\label{Algorithm 2}
\end{algorithm}

In the stochastic variational algorithm (Algorithm 2), updates for $\kappa_{dd'}$ and $\nu_{dd'}$ are cycled until convergence for each $(d,d') \in \mathcal{P}$ so that local parameters are optimized at the current global parameters. The global parameters are then updated using stochastic gradient ascent. For a parameter $\lambda_i$, we consider an update:
\begin{equation} \label{S1}
\lambda_i \leftarrow \lambda_i + s_t \; \tilde{\nabla}_{\lambda_i} \mathcal{L}^*,
\end{equation}
where $s_t$ denotes a small step taken in the direction of $\tilde{\nabla}_{\lambda_i} \mathcal{L}^*$ (natural gradient of $\mathcal{L}^*$ with respect to $\lambda_i$). In variational Bayes and nonconjugate variational message passing, the natural gradient \citep{Amari1998} is $\tilde{\nabla}_{\lambda_i} \mathcal{L}^* = \hat{\lambda}_i - \lambda_i$, where $\hat{\lambda}_i$ is the optimal update of $\lambda_i$. See \ref{suppA} for details. Hence the update in \eqref{S1} can be written as 
\begin{equation*}
\lambda_i \leftarrow (1 - s_t) \lambda_i + s_t  \hat{\lambda}_i .
\end{equation*}
In stochastic gradient ascent, we replace the true natural gradients with unbiased estimates. For convergence, the step sizes should satisfy the conditions $s_t \rightarrow 0$, $\sum_{t=0}^\infty s_t = \infty$ and $\sum_{t=0}^\infty s_t^2 < \infty$.

Using our proposed sampling scheme, unbiased estimates of the true natural gradients can be computed via the Horvitz-Thompson estimator \citep[see e.g.][]{Kolaczyk2010}. Let $\mathcal{P}_{d \Cdot} = \{(l,l') \in \mathcal{P} | l=d\}$ and $\mathcal{P}_{\Cdot d} = \{(l,l') \in \mathcal{P} | l'=d\}$ denote samples in $\mathcal{P}$ lying in the $d$th row and column of the adjacency matrix respectively. We have, for example, the following unbiased estimates,
\begin{equation}\label{HT estimators}
\sum_{(l,l') \in \mathcal{P}_{d \Cdot}} \negthickspace \frac{\kappa_{ll'}}{\pi_{ll'}}  \approx  \sum_{d' \neq d} \kappa_{dd'} \quad \text{and} \quad
\frac{D}{|\mathcal{S}|}\negthinspace \sum_{\substack{(l,l') \in \mathcal{P}_{\Cdot d'} \\ d' \in \mathcal{S}} } \negthickspace \negthickspace \negthickspace \frac{\kappa_{ll'i}\nu_{ll'j}}{\pi_{ll'}} \approx \sum_{(d,d')} \kappa_{dd'i}\nu_{dd'j}.
\end{equation}
The estimator on the right arises from a two-stage cluster sampling scheme which involves sampling the columns ($d' \in \mathcal{S}$) in the first stage and then sampling document pairs from these columns in the second stage. We show that this estimator is unbiased in the \ref{suppA}. 

The stochastic variational algorithm for our model is outlined in Algorithm 2. Implementation details are given in the \ref{suppA}.

\section{Comparison with alternative approaches} \label{Comparison}
We compare our model with the most representative peer methods, RTM and Pairwise-Link-LDA. As a baseline for comparing methods which integrate the modeling of text and links, we also consider ``LDA + Regression", which involves fitting an LDA model to the documents followed by a logistic regression model to the links. The covariates corresponding to the observation $y_{dd'}$ are $\bar{\gamma}_d \odot \bar{\gamma}_{d'}$, where $\odot$ denotes the Hadamard product and the $k$th element of $\bar{\gamma}_d$ is $\gamma_{dk} / \sum_l \gamma_{dl}$. This approach models text and links separately and information from topic modeling is not utilized in the link structure. The RTM accounts for both text and links structure. However, it assumes a symmetric probability function and considers a diagonal weight matrix that allows only within topic interactions. In addition, RTM only models observed links and does not deal explicitly with the imbalance between links and nonlinks. We consider the RTM with an exponential link probability function. Pairwise-Link-LDA combines LDA and MMSB. Comparing Pairwise-Link-LDA with LMV illustrates the importance of the visibility measure in link prediction. 

All models are estimated using variational methods and the code for these algorithms are reproduced in {\ttfamily R} by ourselves. For LDA + Regression, RTM and Pairwise-Link-LDA, we have tried to follow the implementations suggested by the original authors as closely as possible. Variational parameters in LMV, RTM and Pairwise-Link-LDA are initialized using the fitted LDA and the same priors are used across all models. Details on priors and stopping criteria are given in the \ref{suppA}.

\section{Prediction and article recommendations} \label{citation prediction}
We discuss how our model can be used to predict links for new documents assuming knowledge only of the text. An important application of the predictive probabilities is in recommending scientific articles, for instance, to researchers searching for information on certain research topics or who are preparing manuscripts and looking for relevant articles to cite. Using a short paragraph of text or even just keywords (as text of the ``new document"), predictive probabilities of links to documents in the training set can be computed and used as a means to rank documents and construct recommendation lists. As our model captures both inter- and intra-topic citation probabilities and estimates the visibilities of individual articles (which are adjusted for field variation in citation practices), it can identify relevant articles which are multi-topic or of high visibility but coming from topics with low citation rates. \cite{Wang2011b} and \cite{Gopalan2014} consider recommendation of scientific articles to readers of online archives based on article content and reader preferences. They do not consider citations amongst articles and make use of collaborative filtering.
 
The predictive probability of a link to a document in the training set can be computed as follows. First, we fit the LMV model to documents in the training set. Then we perform variational inference on the new document $d$ to obtain its topic proportions \citep[see e.g.][]{Nallapati2008}. That is, we iterate till convergence the updates:
\begin{enumerate}
\item $\gamma_d \leftarrow \alpha +  \sum_n \phi_{dn} $,
\item $\phi_{dnk} \propto \exp\left\{  \psi(\gamma_{dk})-\psi\left(\sum_k \gamma_{dk}\right)+ \sum_v w_{dnv}  \left[\psi(\lambda_{kv})-\psi\left(\sum_v \lambda_{kv}\right)\right]  \right\} $ \\ for $n=1, \dots, N_d$, $k=1, \dots, K$,
\end{enumerate}
where $\lambda$ is obtained from the fitted model. The first update is similar to step 3 of Algorithm~1. In this case, we do not assume knowledge of the links of the new document. Hence parameters associated with the links are absent. Let $w_d$ and $w_T$ denote the words of the new document $d$ and the training set respectively and let $y_T$ denote links within the training set. Approximating the true posterior by the variational approximation $q(\Theta)$, the posterior predictive probability that $d$ will cite any document $d'$ in the training set is
\begin{equation}\label{prediction}
\begin{aligned}
p(y_{dd'}=1| w_d, w_T, y_T) 
&\approx \text{E}_q(\tau_{d'})\text{E}_q( \theta_{d})^T \text{E}_q(B) \text{E}_q(\theta_{d'})  \\
&=\frac{g_{d'}}{g_{d'}+h_{d'}} \bigg(\frac{\gamma_d}{\sum_{k=1}^K \gamma_{dk}} \bigg)^T \frac{a}{a+b} \; \frac{\gamma_{d'}}{\sum_{k=1}^K \gamma_{d'k}}.
\end{aligned}
\end{equation}

\subsection{Predictive rank} \label{S predictive rank}
In the examples, we compute the average predictive rank of held-out documents, following \cite{Chang2010}, as a way to evaluate the fit between considered models and the data. The predictive rank captures a model's ability to predict documents that a test document will cite given only its words. To compute the predictive rank of a test document for model $\mathcal{M}$, first fit model $\mathcal{M}$ to documents in the training set. Using the fitted topics, obtain the topic proportions of the test document using just its words as described above. Compute the posterior predictive probability that the test document will cite each document in the training set (use \eqref{prediction} for LMV) and rank the documents according to this probability. The predictive rank is the average rank of the documents which the held-out document actually did cite. Lower predictive rank indicates a better fit, and it also implies that the articles that were actually cited are placed closer to the top of a recommendation list for a test article.

\section{Applications} \label{examples}
We apply our methods to two real datasets. To assist understanding of our proposed model and to evaluate Algorithms 1 and 2, a simulation study is also provided in the \ref{suppA}. In this study, we generate datasets from the LMV and demonstrate that Algorithms 1 and 2 are able to recover the structure of the blockmodel $B$ as well as the visibility of each document. We also show that our model performs significantly better than LDA+Regression, RTM and Pairwise-Link-LDA in link predictions. Predictive results from Algorithms 1 and 2 are very close and our subsampling strategy helps to reduce computation times. 

In the following, the blockmodels, visibilities, topic proportions and topic assignments of documents are estimated using the posterior means of corresponding variational approximations. For each citation, a hard assignment of topics to the citing and cited documents is obtained by taking the positions of the maximum elements of $\kappa_{dd'}$ and $\nu_{dd'}$ respectively. For instance, if $\kappa_{dd'} = [0.8, 0.2, 0, \dots, 0]$ and $\nu_{dd'}=[0.3, 0.7, 0 \dots, 0]$, we interpret the citation as being from topic 1 to 2. For visualization of fitted topics, we order terms in the vocabulary using the term-score \citep{Blei2009},
\begin{equation*}
\text{term-score}_{kv} = \bar{\lambda}_{kv} \log \Bigg\{ \frac{\bar{\lambda}_{kv}}{ \big(\prod\nolimits_{k=1}^K \bar{\lambda}_{kv} \big)^ \frac{1}{K}} \Bigg\},
\end{equation*}
where $\bar{\lambda}_{kv}=\lambda_{kv}/\sum_l \lambda_{kl}$ denotes the posterior mean probability of the $v$th term appearing in the $k$th topic. The second part of the expression downweights terms that have high probability of appearing in all topics. This term-score is inspired by the TFIDF term score used in information retrieval \citep{Baeza1999}. Further discussion on ways to examine the quality of uncovered topics can be found in the \ref{suppA}. 

We denote LDA + Regression and Pairwise-Link-LDA by LDA+Reg and PLLDA respectively. Methods with subsampling have a ``S" added at the end, e.g. LMVS denotes LMV with subsampling. If publication times are taken into account, we add a ``$(t)$" at the end.

\subsection{Cora dataset} \label{Cora_eg}
The Cora dataset from the {\ttfamily R} package lda \citep{Chang2012} has 2410 documents, 4356 links and a vocabulary of 2961 terms.  This dataset is relatively small and it allows us to compare the predictive performance and CPU times of Algorithms 1 and 2.  We randomly divide the dataset into five folds; each fold is used in turn as a test set and the remaining folds are used for training. During training, only documents in the training set and links within them are used. 
We investigate the predictive performance of different models for number of topics $K$ ranging from 5 to 13. For Algorithm 2, we set the minibatch size as 200. 
 
\subsubsection{Predictive performance and computation times} 
\begin{figure}[tb!]
\centering
\begin{tabular}{cc}
Predictive Rank & CPU Runtime \\
\hspace*{-2mm}\includegraphics[width=0.5\textwidth]{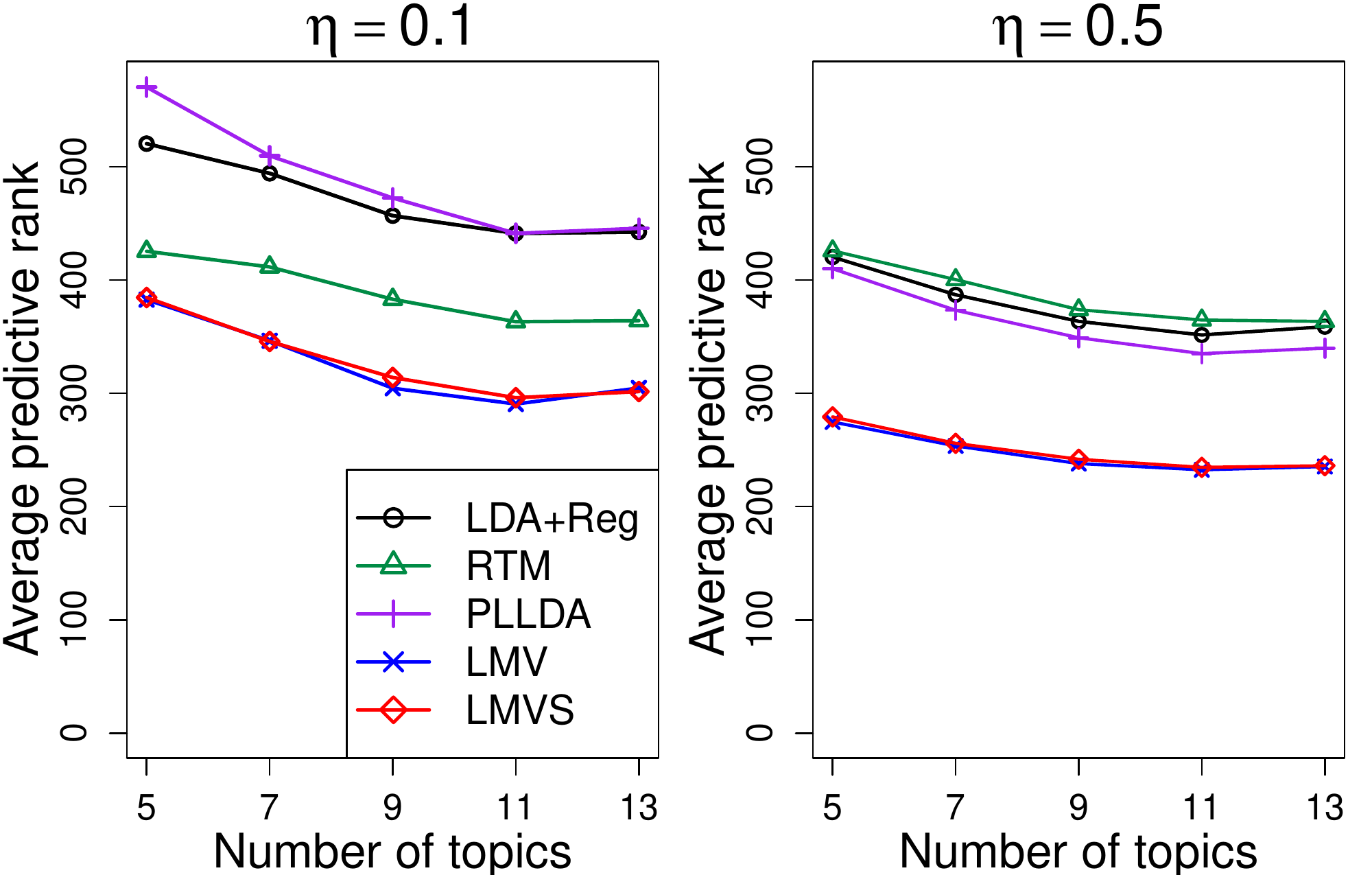} & \hspace{-3mm}\includegraphics[width=0.5\textwidth]{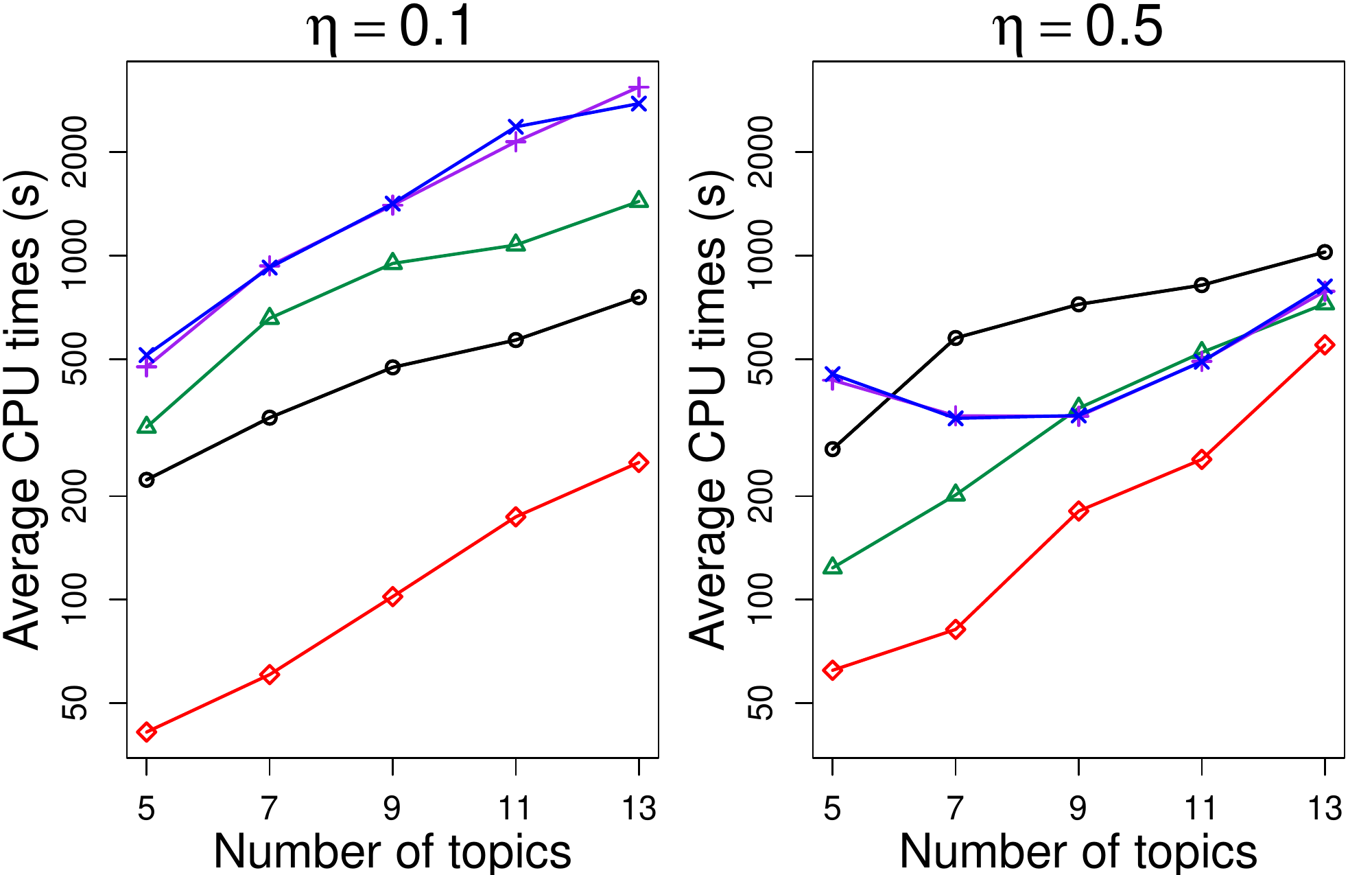}
\end{tabular}
\caption {\label{Coraranktimes} Cora: Average predictive ranks and CPU times in seconds for different methods.}
\end{figure}
The average predictive ranks and CPU times of different approaches are shown in Figure~\ref{Coraranktimes}. We consider the hyperparameter $\eta$, which controls the concentration of the topic distribution, to be 0.1 or 0.5 (if $\eta$ is small, the probability distribution on the vocabulary will be concentrated on a small number of terms). Except for the RTM, predictive performance of all other methods are better when $\eta$ is 0.5 as compared to 0.1. LMV achieved significantly better predictive performance than the other models and attained 60--76\% improvement in predictive rank over baseline\footnote{Predictive rank computed by random guessing is $\frac{n+1}{2}$ where $n$ is the number of documents in the training set.}. Predictive performance of Algorithm 2 is close to that of Algorithm 1 even with subsampling and computation times are reduced, particularly when $\eta=0.1$. For large datasets, Algorithm 2 presents an avenue for overcoming computational and memory constraints while maintaining the same level of predictive performance. Predictive performance for the LMV stabilizes at around 9--11 topics when $\eta=0.5$. In the following, we concentrate on the fitted models corresponding to $K=9$ and $\eta=0.5$ for one of the folds. 

\subsubsection{Evaluating accuracy of Algorithm 2} 
\begin{figure}[tb!]
\centering
\includegraphics[width=0.75\textwidth]{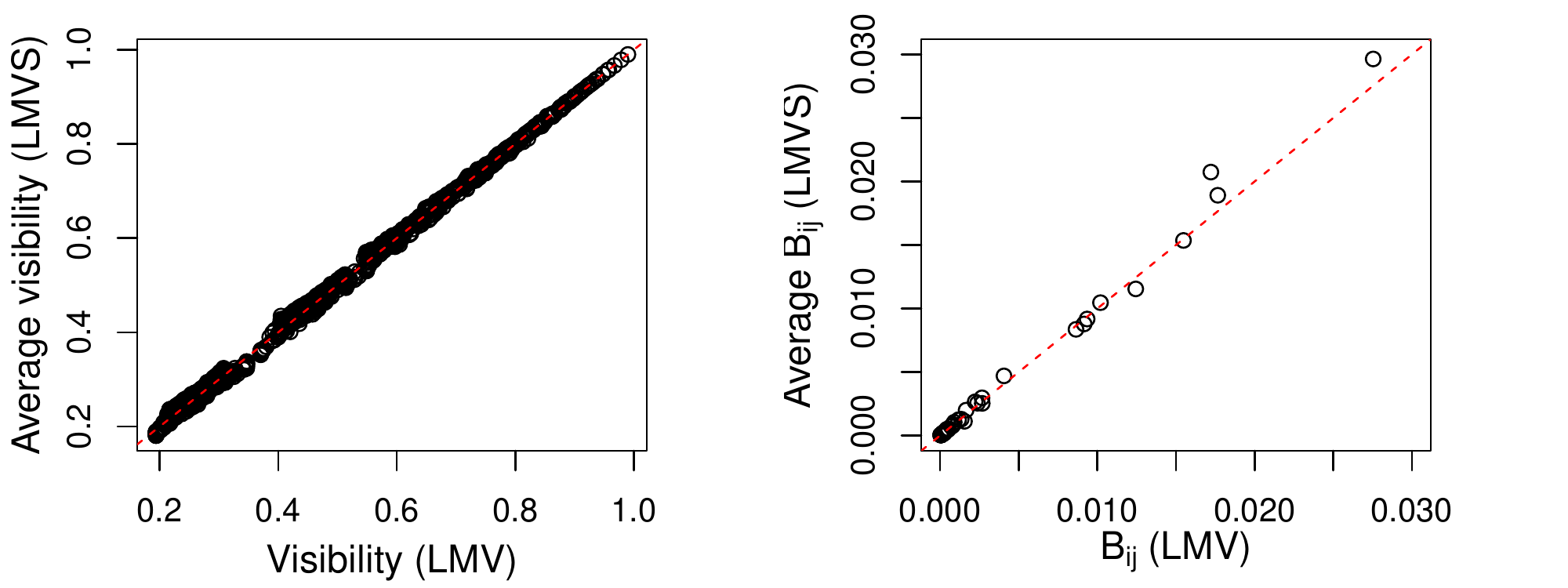}
\caption {\label{CoraLMVLMVS} Cora: Visibilities (left) and blockmodel elements (right) estimated by LMVS and averaged over 50 runs against corresponding qualities estimated by LMV. Points lying close to the dotted line ($y=x$) indicate good agreement between LMV and LMVS.}
\end{figure}
Repeating the LMVS (Algorithm 2) runs 50 times, we compute the average visibilities and blockmodel over these 50 runs and plot these quantities against corresponding values estimated by LMV (Algorithm 1) in Figure \ref{CoraLMVLMVS}. There is very good agreement between the visibilities and blockdiagonal elements estimated by Algorithms 1 and 2. For the left plot, there is greater variation near zero while the biggest two values appear to be slightly overestimated by LMVS in the right plot.

\subsubsection{Visibilities of individual articles} 
\begin{figure}[tb!]
\centering
\includegraphics[width=0.7\textwidth]{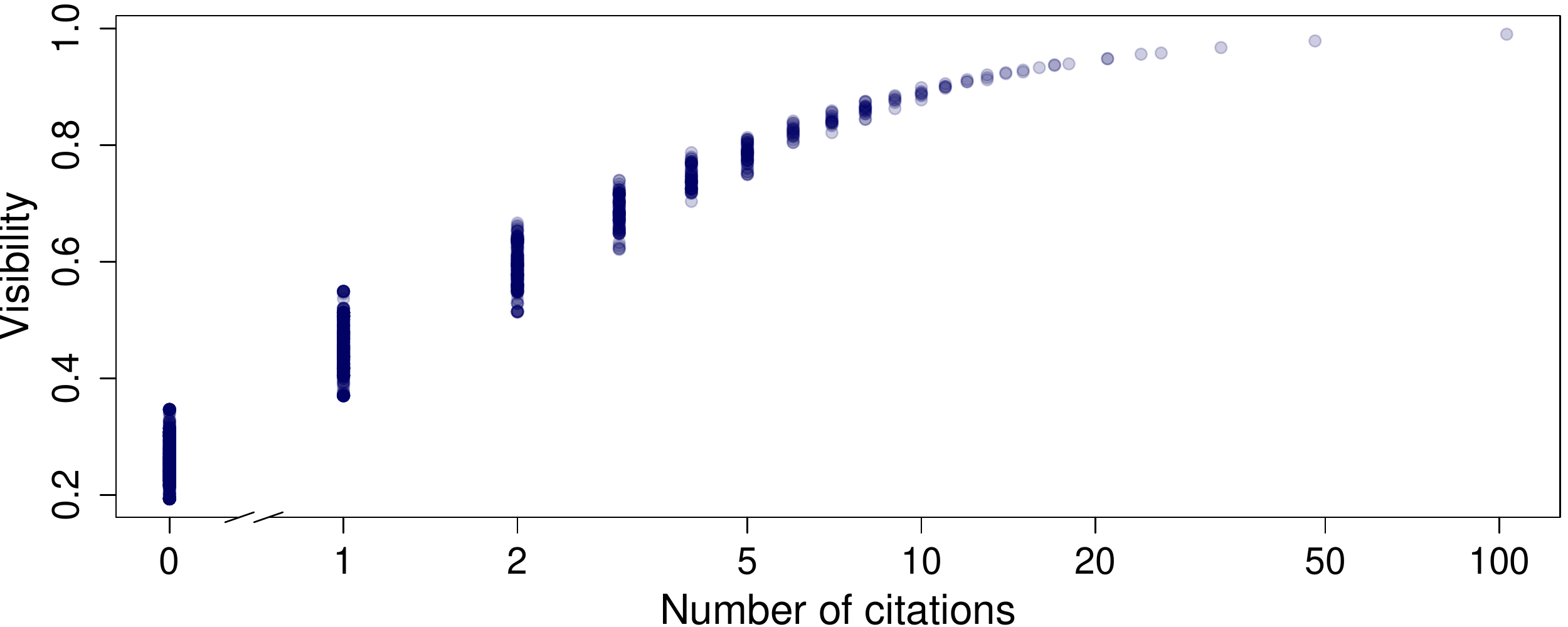} 
\caption {\label{Coralq} Cora: Visibility against number of citations for each document in training set.}
\end{figure}

Figure \ref{Coralq} plots the visibility of documents in the training set estimated by Algorithm 1 against their citation counts. There is a general trend of visibility increasing with citation counts. Hence, the topic-adjusted visibility metric is capturing some degree of popularity. However, the relationship between visibility and citation counts is not monotone; a higher citation count does not necessarily imply a higher measure of visibility. The visibility metric $\tau_{d'}$ captures a complex mix of attributes of document $d'$ that accounts for the variation in citation probability among documents with similar topic proportions. These include but are not limited to popularity.

To illustrate how the incorporation of visibilities improves predictive performance, let us consider as an example, the test document {\it ``Some extensions of the k-means algorithm for image segmentation and pattern classification"}. This article cited two documents from the training set: (1) {\it A Theory of Networks for Approximation and Learning} and (2)  {\it Self-Organization and Associative Memory}. The citations, estimated visibility and predictive ranks of these documents are given in Table \ref{Coracited}.
\begin{table}
\caption{Cora: Citations, visibility and predictive ranks of two cited documents.}
\label{Coracited}
\centering
\small
\begin{tabular}{cccccccc} 
\hline
Index  & Citations & $\hat{\tau}$ & LDA+Reg  &  RTM &  PLLDA & LMV & LMVS \\ \hline
1 & 26 & 0.96 & 98  & 108  & 120  & 10 & 9\\ 
2 & 48 & 0.98 & 385 & 1082 & 336 & 53 & 61\\ \hline
\end{tabular}
\vspace{2mm}
\end{table}
The predictive ranks assigned by LMV and LMVS are significantly lower than the other methods. Interestingly, if visibility is not taken into account, the ranking by LMV will be similar to Pairwise-Link-LDA (120 and 357 for documents indexed 1 and 2 in order). However, factoring in the visibilities of these documents, which are much higher than the average of 0.42, improves predictive ranks tremendously. On the average, LMV improves  predictive performance by more than 30\% over RTM, PLLDA and LDA+Reg when $\eta=0.5$. This provides an indication of the favorable performance of LMV in article recommendation.

\subsubsection{Citation behavior among fitted topics} 
\begin{figure}[tb!]
\begin{minipage}{0.555\textwidth}
\centering
 \includegraphics[width=\textwidth]{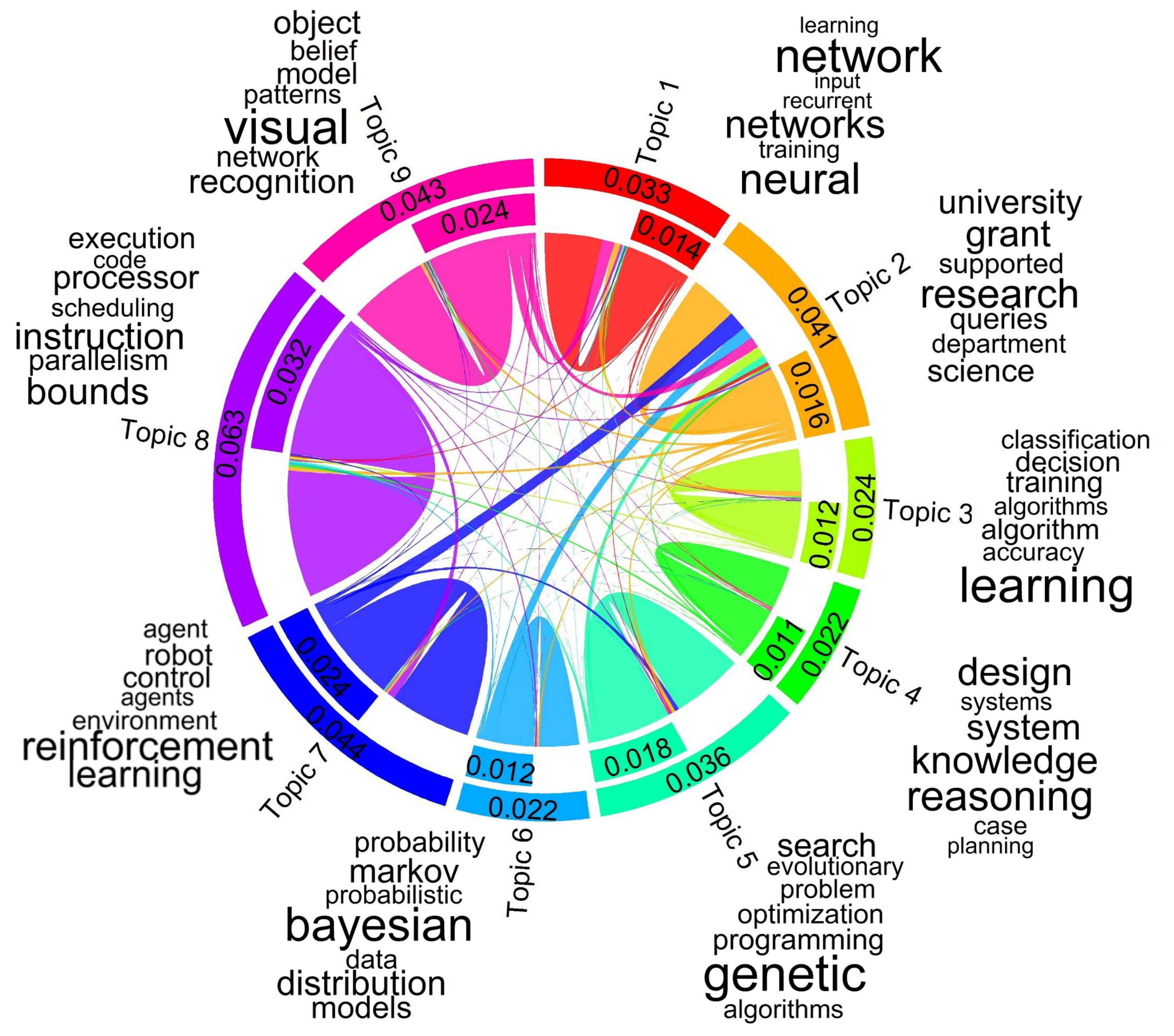} \text{(a) \it Blockmodel $B$ and top words of each topic}
\end{minipage}
\begin{minipage}{0.435\textwidth}
\centering
 \includegraphics[width=\textwidth]{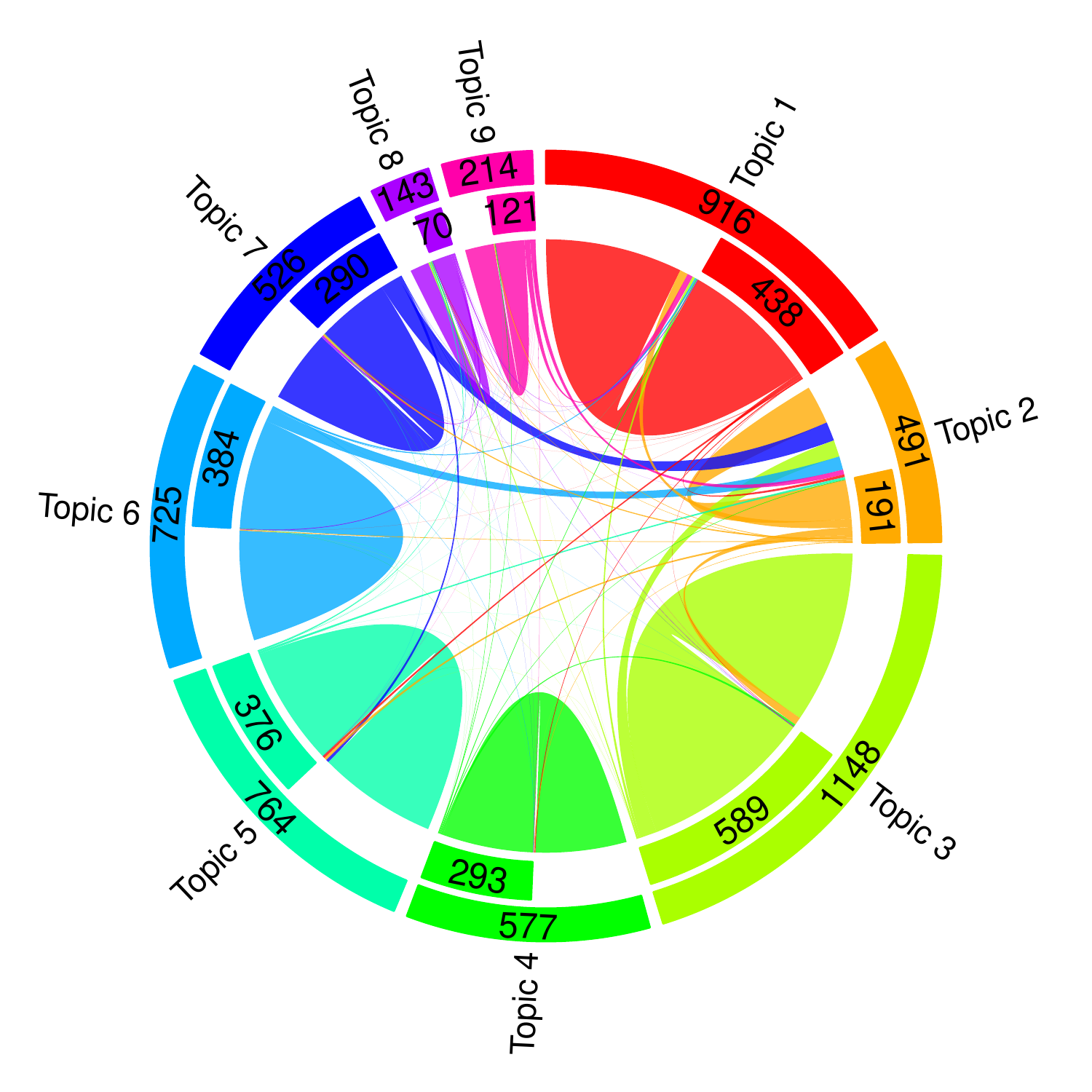} \text{(b) \it Citation activity} 
\end{minipage}
\caption{(a) shows the citation patterns between different topics and the topic-to-topic citation tendencies. Each element $B_{ij}$ is represented by a belt from topic $i$ to $j$ and the width of the belt is proportional to $B_{ij}$. The value on the inner arc of topic $i$ represents the probability that topic $i$ will cite any topic while the value on the outer arc represents the total probability that topic $i$ will cite or be cited by any topic. The top 7 words of each topic are displayed in word clouds and the font size is proportional to the term-score. (b) shows the actual citation activity among different topics. The width of each belt is proportional to the number of citations between the topics connected by the belt. The value on the inner arc is the total number of citations originating from a topic while the value on the outer arc is the total number of citations coming from and going to that topic. The direction of the belts can be inferred from the presence (at the origin) and absence (at the destination) of the inner arcs.}
\label{CoraBactivity}
\end{figure}
Figure \ref{CoraBactivity} provides a visualization of the LMV fitted using Algorithm 1, where (a) shows the estimated blockmodel $B$ and top words of each topic and (b) shows the citation activity in the training set. Figure 6(b) is based on raw citation counts and tends to be dominated by topics with high citation frequency or large number of publications. On the other hand, Figure 6(a) shows the citation probabilities within/between topics. It provides a more structured and unbiased understanding of the citation tendencies within/between topics. From Figure \ref{CoraBactivity}(a), topic 2 tends to be cited strongly by topics 3, 5--7 and 9. It also tends to cite nearly all other topics besides itself. Topic 9 tends to cite topics 1 and 2, and topic 8 has the highest tendency to cite documents within its own topic. Information such as these are helpful in understanding how citation behavior varies from one scientific area to another, and even among different research fields within a discipline.

\subsection{KDD high energy physics dataset} \label{KDD_eg}
The high energy physics (HEP) dataset for the KDD Cup 2003 competition \citep[see][]{Gehrke2003} contains 29,555 papers added to arXiv from 1992--2003 and is available at \url{http://www.cs.cornell.edu/projects/kddcup/index.html}. We consider the abstracts of 25,224 papers added between 1992--2001 and the 271,838 citations among them as the training set. We use 4064 papers which have cited at least one paper from the training set and were added between 2002--2003 as the test set. There are 60,914 citations from the test set to the training set. The vocabulary consist of 7211 terms after stemming and removal of stop words and infrequent terms. We have the dates when the papers were published online at the library of the Stanford Linear Acceleration Center (SLAC), as well as the year and month in which the papers were added on arXiv. As some papers were first published on SLAC and later added on arXiv or vice versa, we use the earlier of the two dates as the date when a paper is first available. Memory constraints render running the algorithms for Pairwise-Link-LDA and LMV in batch mode infeasible. Hence, we only use Algorithm 2 and Pairwise-Link-LDA is also implemented using our proposed subsampling strategy. Minibatch size of 2000 was used.

\subsubsection{Predictive performance and computation times} 
\begin{figure}[tb!]
\centering
\includegraphics[width=0.85\textwidth]{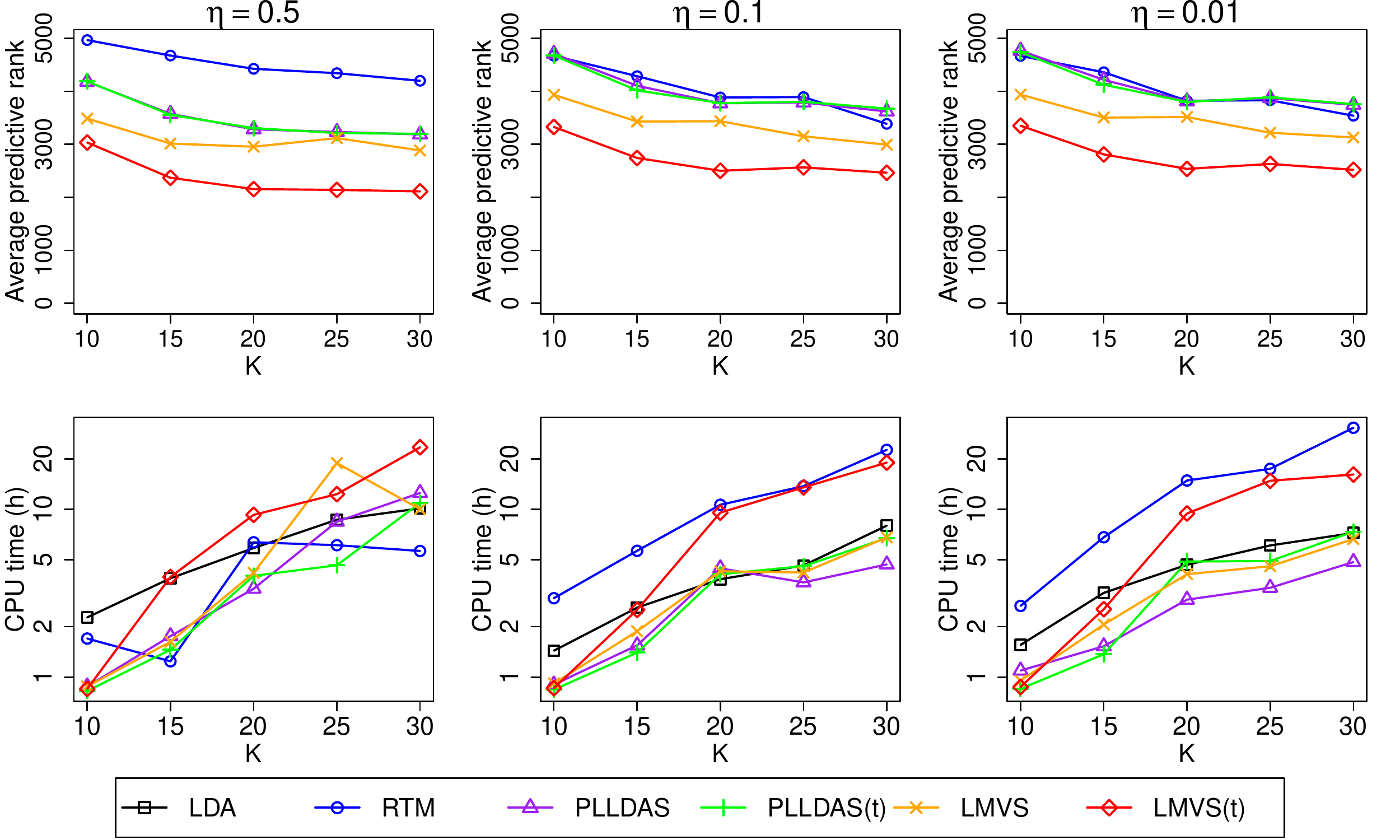}
\caption {\label{KDDranktimes} KDD: Average predictive ranks (first row) and CPU times in hours (second row) of different approaches. The columns correspond to $\eta \in \{0.5, 0.1, 0.01 \}$ from left to right.}
\end{figure}
Figure \ref{KDDranktimes} shows the average predictive ranks and CPU times\footnote{LDA was run in batch mode and CPU times are for model fitting only. We do not compute predictive ranks for LDA+Reg as logistic regression for this dataset is prohibitively expensive. Note that LDA may also be implemented with stochastic variational inference.} of different approaches for number of topics $K$ ranging from 10 to 30 and $\eta \in \{0.5, 0.1, 0.01 \}$. LMV provides better predictive performance than Pairwise-Link-LDA and RTM for all $\eta$ and $K$, and taking into account publication times yields significantly better predictions. This is likely due to more accurate estimation of the visibilities as documents published at later dates are not penalized for not being cited by documents published before them. LMVS(t) achieved 73--83\% improvement over baseline and optimal predictive performance at $K=20$ and $\eta=0.5$. In the following, we consider the model fitted by LMVS(t) when $K=20$ and $\eta=0.5$.

\subsubsection{Interpreting citation trends in HEP} 
\begin{figure}[tb!]
\begin{minipage}{0.48\textwidth}
\centering
 \includegraphics[width=\textwidth]{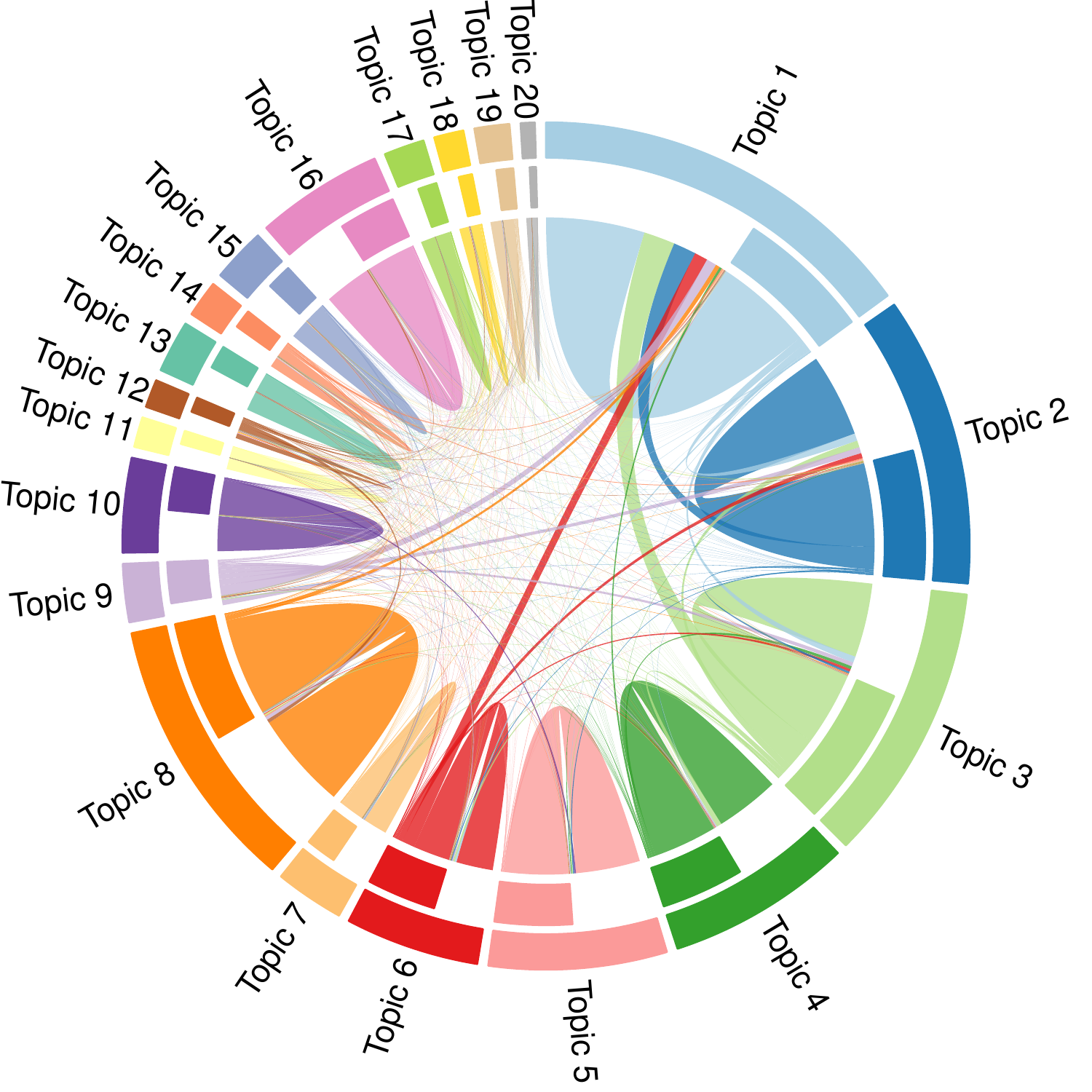}  \caption{Blockmodel $B$}  \label{KDDB}
\end{minipage}
\begin{minipage}{0.48\textwidth}
\centering
 \includegraphics[width=\textwidth]{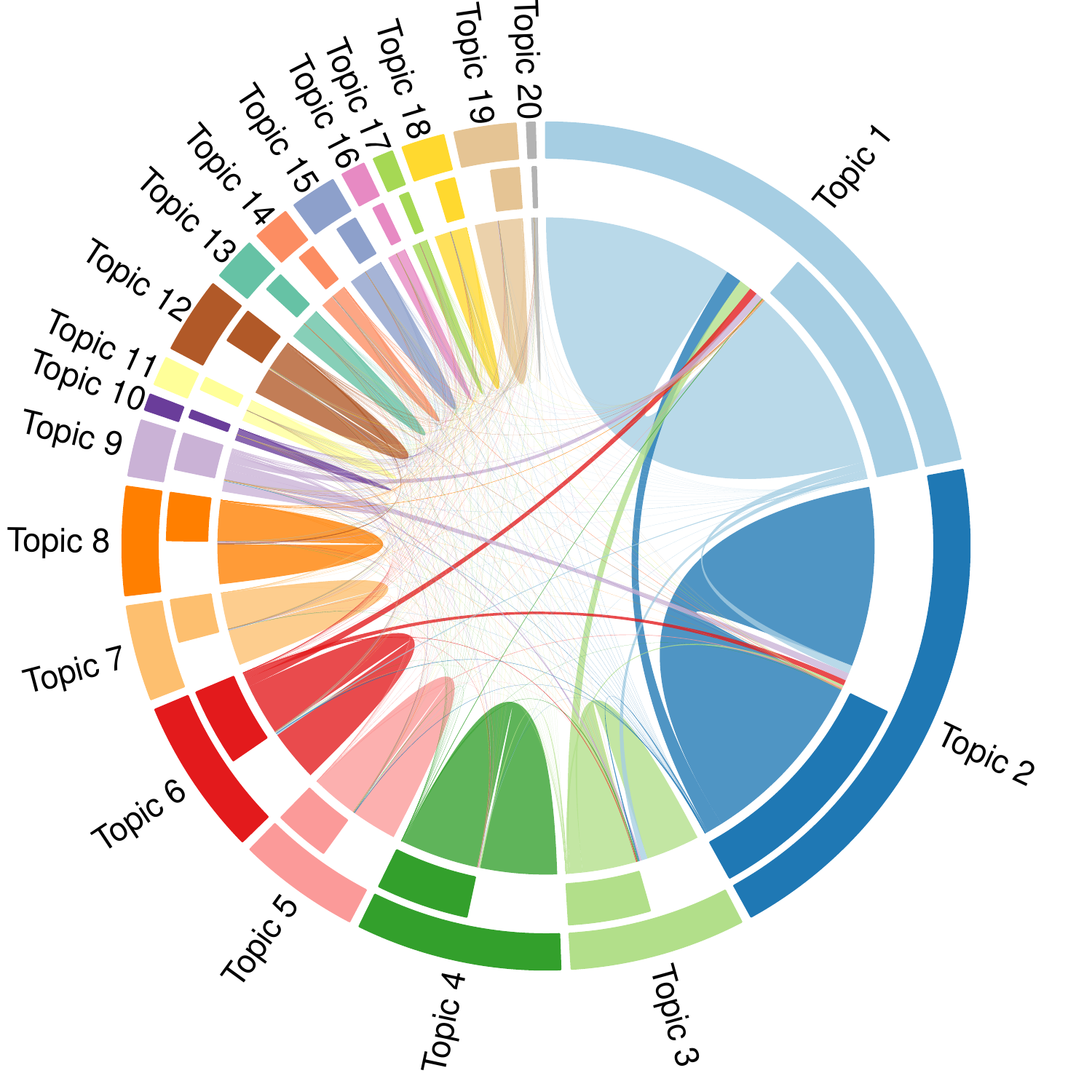} \caption{Citation activity} \label{KDDactivity}
\end{minipage}
\end{figure}

Figures \ref{KDDB} and \ref{KDDactivity} provide visualizations of the estimated blockmodel and the citation activity in the training set respectively. These plots may be read as in Figure \ref{CoraBactivity}. The blockmodel in Figure \ref{KDDB} indicates high probability of within topic citation generally while across topic citation is much weaker in comparison. Figure \ref{KDDactivity} indicates that a large proportion of citations in the corpus occurred within topics 1--2.

\begin{figure}[tb!]
\centering
\includegraphics[width=0.8\textwidth]{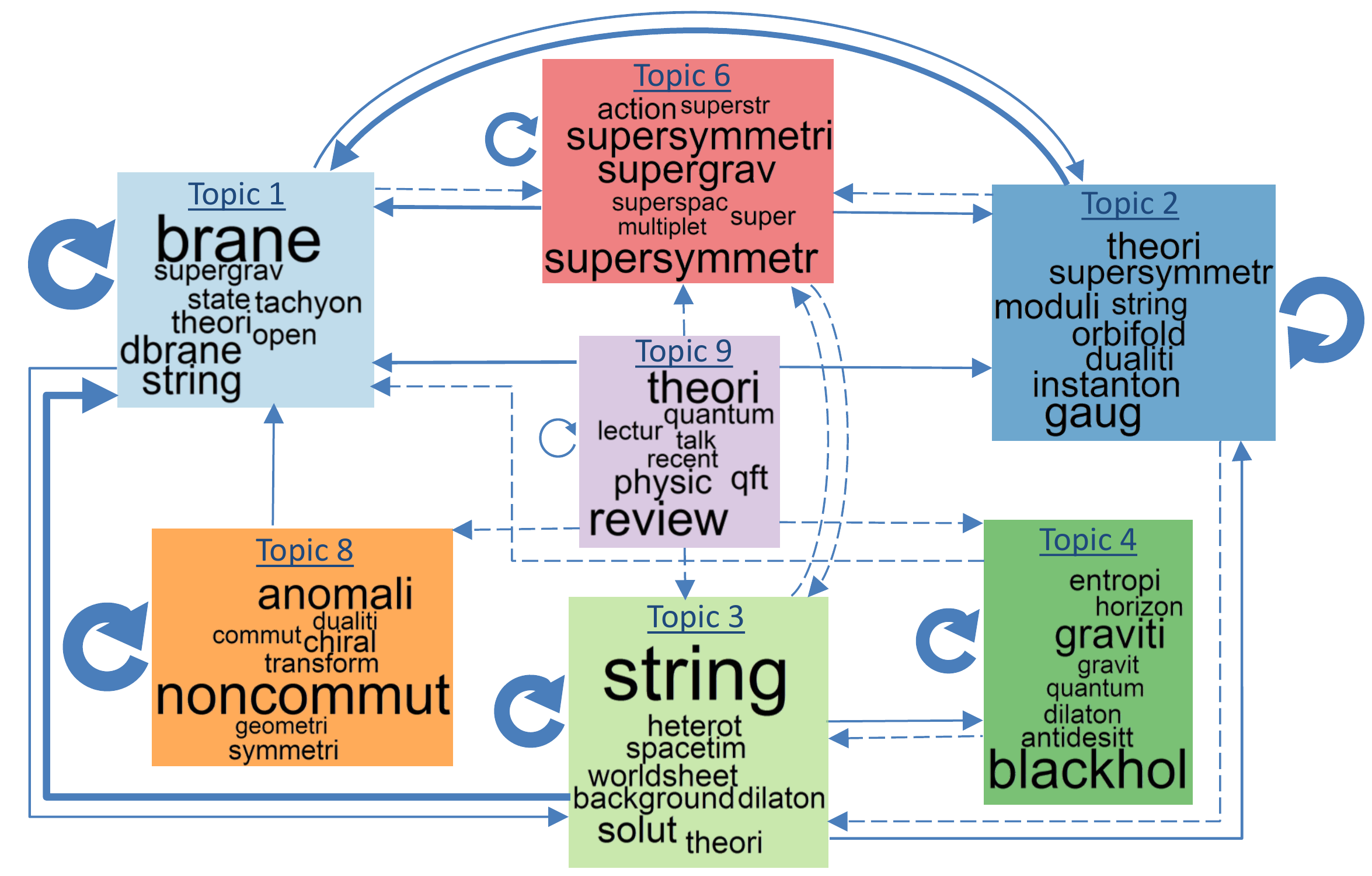}
\caption {\label{KDDtp} KDD: Visualization of citation patterns for topics 1--4, 6 and 8--9. Width of arrows are proportional to the citation strengths in the estimated blockmodel. Only elements greater than 0.001 are visualized. Dashed arrows represent elements less than 0.005. Top 8 words of each topic are shown. Font size within each topic is proportional to the term-score.}
\end{figure}
Next, we focus on topics 1--4, 6 and 8--9 where interconnectivity is higher among them. Figure \ref{KDDtp} provides a visualization of the citation patterns between these topics and reveals some interesting trends in the citation landscape of HEP. First, there is a strong tendency for within topic citations while the probability of across topic citations is much weaker. However, the across topic citation relations that do exist are unsurprising as theoretical HEP deals with the fundamental aspects of Particle Physics and there are vast and deep links between these topics. It is also worth noting that certain topics do not have a high tendency to cite each other (e.g. topics 3 and 6) even though there are overlaps in the body of knowledge (e.g. supersymmetries and string theory). A possible reason is that HEP physicists may not always be aware of one another's work (especially amongst the 3 main groups: experimental, phenomenological and theoretical). The articles from these groups have different focus and may constitute different topics.

Topics 1 and 2 are both associated with string theory, which claims that the fundamental objects that make up all matter are strings (like rubber bands) instead of particles \citep[e.g.][]{Gubser2010}. The emphasis of topic 1 is on branes, which are multi-dimensional membranes that generalize the concept of particle (zero-dimensional brane) and string (one dimensional brane) to higher dimensions. Topic 2 is associated with gauge theory and orbifold, which are both related to geometry. Duality, an important concept in string theory, relates branes (topic 1) to gauge theory and supersymmetry (topic 2). Thus, the citation relationship between topics 1 and 2 is expected. From Figure \ref{KDDtp}, these two topics also have a relatively higher probability of receiving citations from other topics. This could be socially driven by the authors' inclination to cite papers from which their respective topics originated. There may also be a tendency for researchers to appeal to popular topics and cite earlier successful theories such as the gauge theory which is the fundamental edifice of high energy particle physics.

The prefix ``super'' is found in topics 1, 2 and 6. It is associated with supersymmetric theories which attempt to unify bosons and fermions under one generalized scheme; by relating fractional spin to integral spins, and finally unifying force and matter particles. Links from topic 6 to topics 1 and 2 are therefore reasonable and expected. Topics 1 and 3 are clearly related, namely: brane, string and gravity. String theory holds the promise to unify gravity to the other fundamental forces in Physics at the expense of introducing more dimensions that needs to be compactified mathematically. In the case of topic 4, research concerning entropy may be cited in relation to the event horizon of the black hole and this constitutes within-topic citations. Mini black holes may also be regarded as a collection of dbranes, resulting in citations from topic 4 to 1. For future investigation, it may be of interest to narrow the study to a certain HEP group (e.g. experimental, phenomenological or theoretical) and a period when a particular topic is in fashion.

\subsubsection{Visibilities of individual articles}
\begin{figure}[tb!]
\centering
\includegraphics[width=0.95\textwidth]{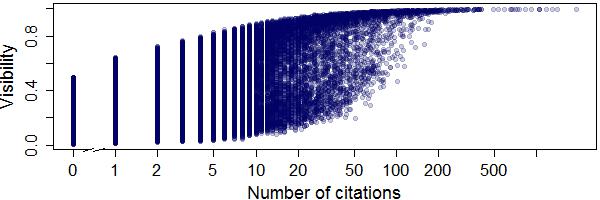}
\caption {\label{KDDlq} KDD: Visibility against number of citations for each document in training set.}
\end{figure}
As in Figure \ref{Coralq}, Figure \ref{KDDlq} plots the estimated visbility of documents in the training set against their citation counts. There is a general trend of visibility increasing with citations as before. However, the estimated visibility for each citation count now vary over a wide range. This range increases as the citation count increases from 0 but starts to gradually decrease as the citation count exceeds 20. It is more evident here than in Figure \ref{Coralq} that visibility correlates with but does not increase monotonically with citations. Hence the visibility metric captures a complex mixture of attributes that includes but goes beyond popularity. 

\subsubsection{Application: article recommendation}\label{S art rec}
\begin{figure}[tb!]
\centering
\includegraphics[width=0.99\textwidth]{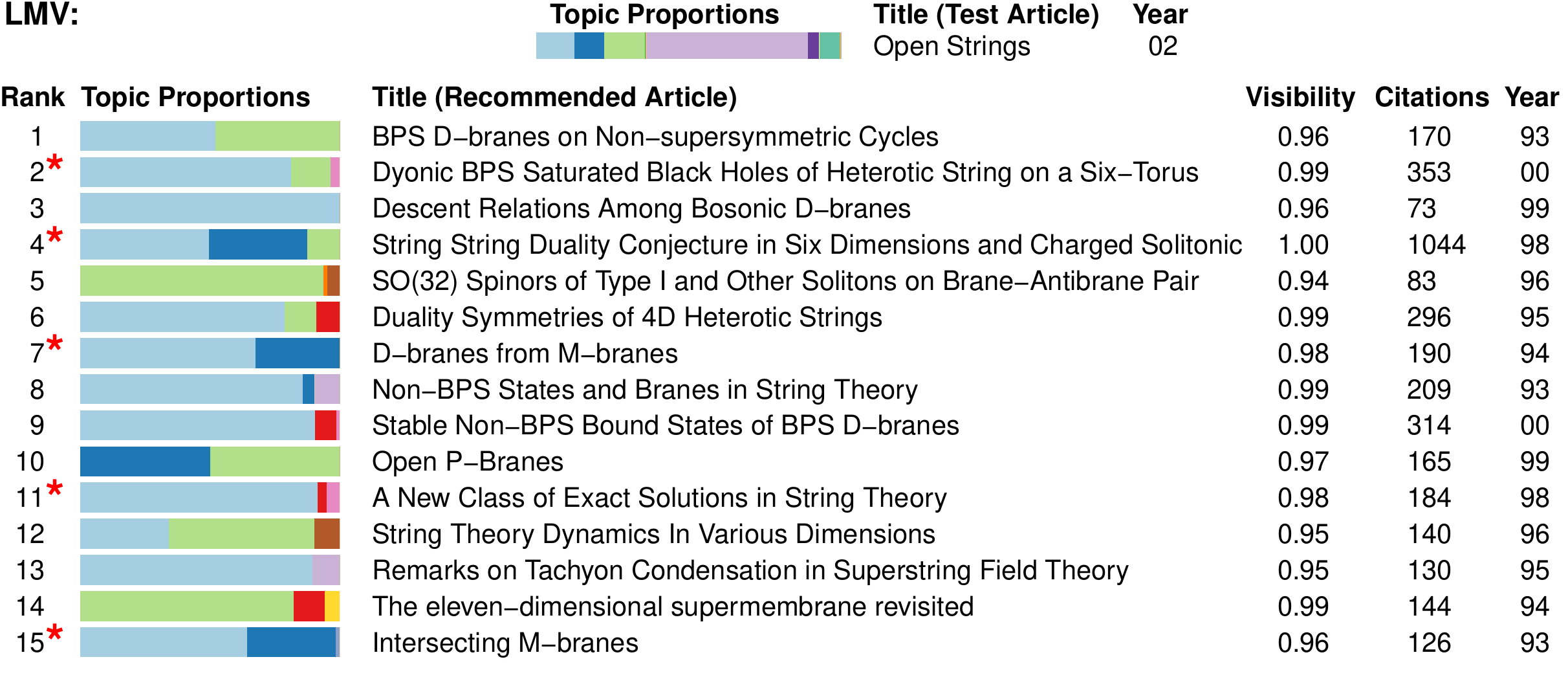}
\includegraphics[width=0.99\textwidth]{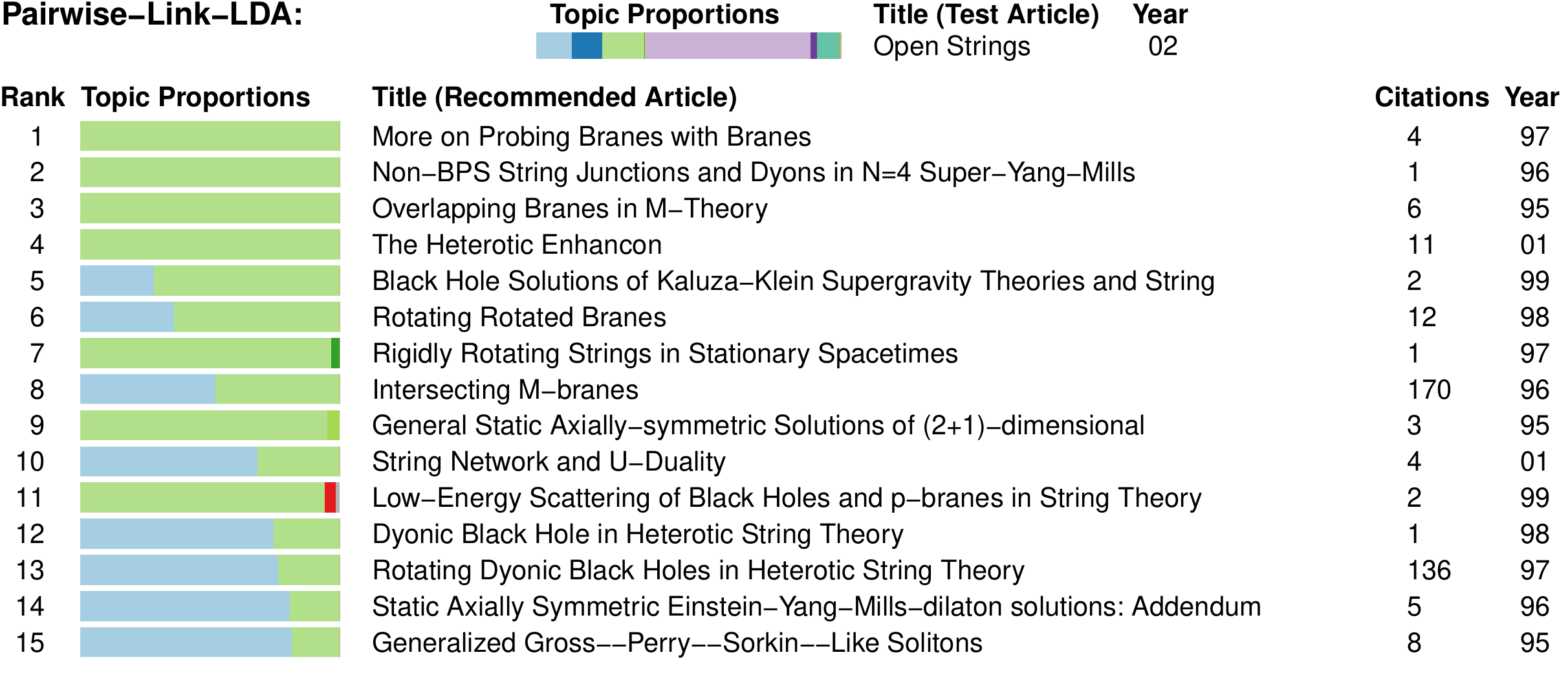} 
\includegraphics[width=0.99\textwidth]{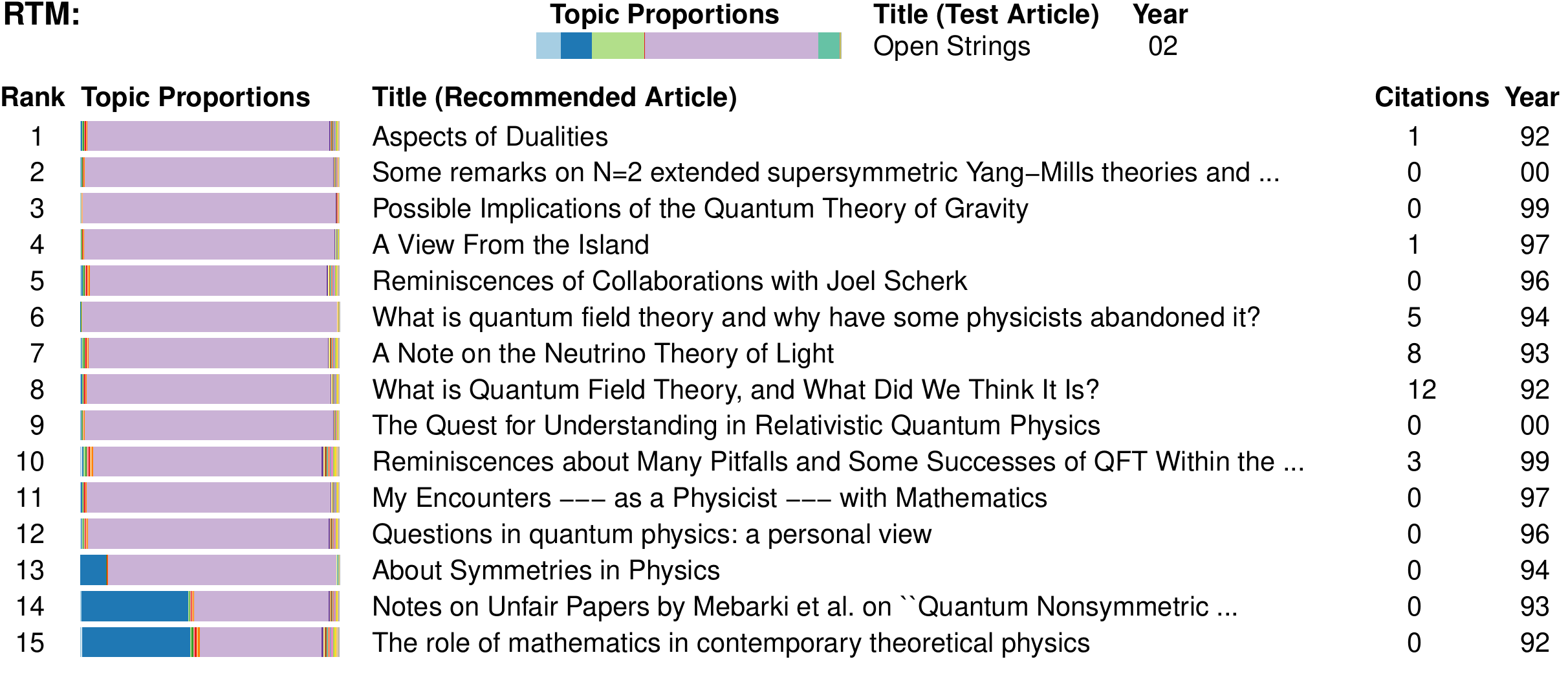}  
\includegraphics[width=0.99\textwidth]{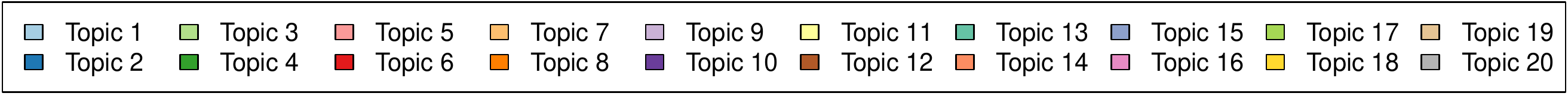}
\vspace*{-3mm}
\caption {\label{KDDartrec} KDD: Top 15  articles recommended by LMV, Pairwise-Link-LDA and RTM. Asterisks indicate articles actually cited by test article. This figure is intended to be read in color (color version available in online publication).}
\end{figure}
We use an example to illustrate the advantages that LMV can provide in article recommendation. Using the abstract of the article ``Open Strings" from the test set as a query, we compute predictive probabilities of links from this article to each document in the training set using LMV, Pairwise-Link-LDA and RTM. In practice, keywords or a short paragraph of text may be used as search query if the abstract or manuscript is not available.  Figure \ref{KDDartrec} lists the top fifteen recommended articles for each approach based on predictive probabilities (articles ranked closer to the top have higher probabilities of a link). Information such as topic proportions (in the form of barplots), title, citation counts, year of publication and visibility metric (for LMV) are also provided. This is a realistic representation of recommendation systems that can be constructed based on our proposed model. 

In this example, five of the fifteen articles (marked by red asterisks) recommended by LMV were actually cited while none recommended by RTM or Pairwise-Link-LDA were cited. Comparing the topic proportions of the test article with those of the recommended articles, we note that RTM tend to recommend articles with high proportions of topic 9. This is likely because RTM only allows within topic interaction and the test article exhibits high proportions of topic 9. On the other hand, Pairwise-Link-LDA and LMV are able to model across topic citation tendencies and the probability of topic 9 citing topics other than itself is quite high (see Figures \ref{KDDB} and \ref{KDDtp}). Hence Pairwise-Link-LDA recommends articles mainly from topics 1 and 3 while articles recommended by LMV exhibit a higher degree of mixing with topics from 1--3 mainly and smaller proportions of some other topics. The articles recommended by LMV tend to have more citations than those recommended by  Pairwise-Link-LDA and RTM. This is because the visibility metric captures a certain degree of popularity as shown in Figure \ref{KDDlq}. However, articles are not ranked based on citations alone; both topic compatibility and article visibility play a part in the ranking. For LMV, there is also a good mix of articles published from 1993--2000. While our model does not explicitly model time from publication, the visibility metric accounts for this to a certain degree in that old articles which have accumulated many citations will high visibility, but so will recent articles which have managed to garner a proportionately smaller number of citations. This example highlights some of the advantages that LMV has to offer for article recommendation such as being able to identify relevant multi-topic articles and taking article visibility into account in rankings.

\subsubsection{Visibility as a topic-adjusted measure}
\begin{figure}[tb!]
\centering
\includegraphics[width=0.85\textwidth]{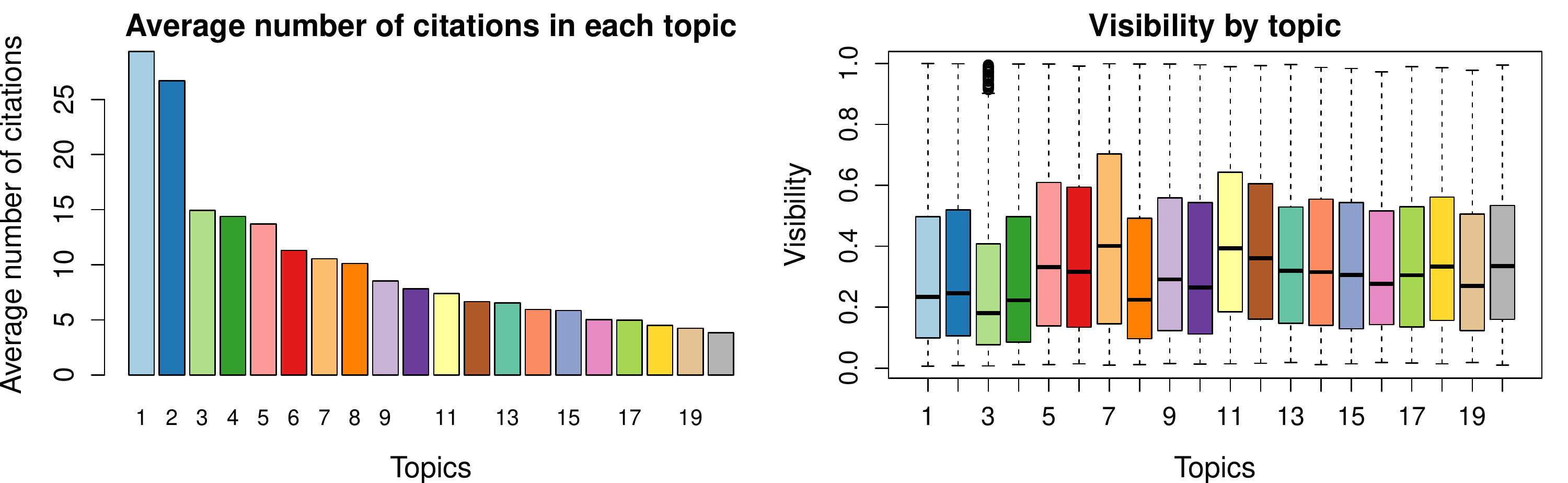} \\
\includegraphics[width=0.85\textwidth]{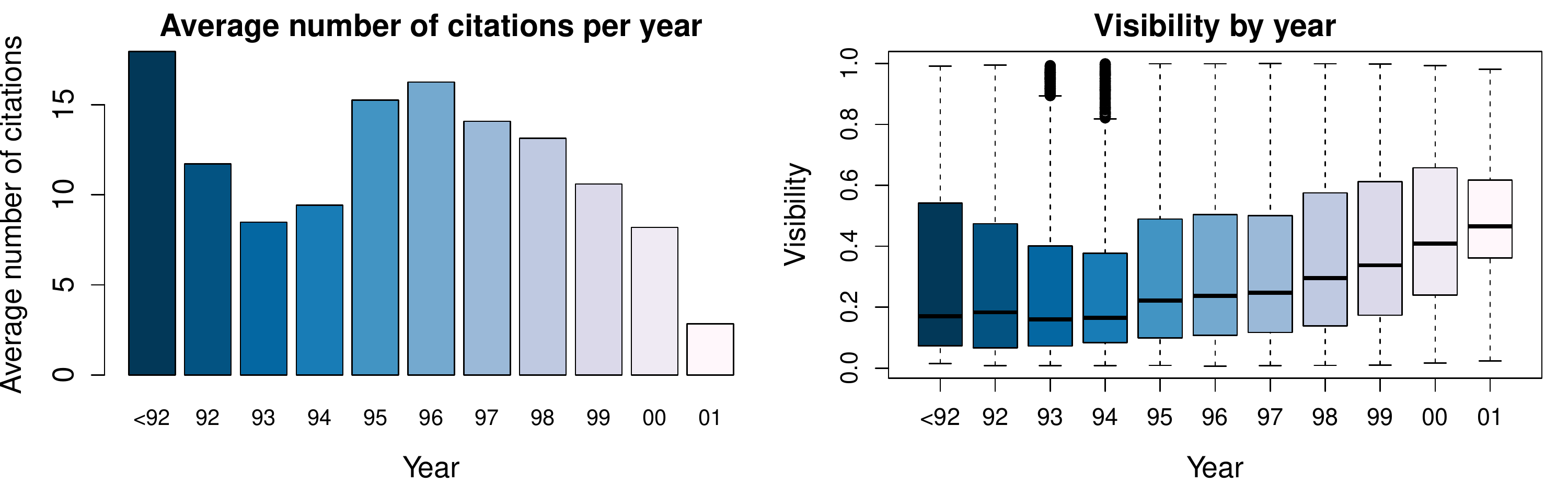}
\caption {\label{KDDvisbilitytopicyear} KDD: First row shows a barplot of the average number of citations in each topic (left) and boxplots of the visibilities of training documents classified by topic (right). Second row shows a barplot of the average number of citations in each year (left) and boxplots of the visibilities of training documents classified by year (right).}
\end{figure}
We examine the behavior of the visibility measure by plotting the estimated visibilities by topic and by year of publication in Figure \ref{KDDvisbilitytopicyear}. We note that the visibilities do not vary radically from one topic to another and are in fact much more comparable across different topics than the average number of citations per article. The boxplots of visibilities by year does not display any systematic correlation with time of publication as well. Interestingly, even as the average number of citations per year is decreasing rapidly for articles published in the later years, the visibilities are not showing any decreasing trend. This could be due to the tendency to cite the newest and latest research.

\begin{figure}[tb!]
\centering
\includegraphics[width=0.9\textwidth]{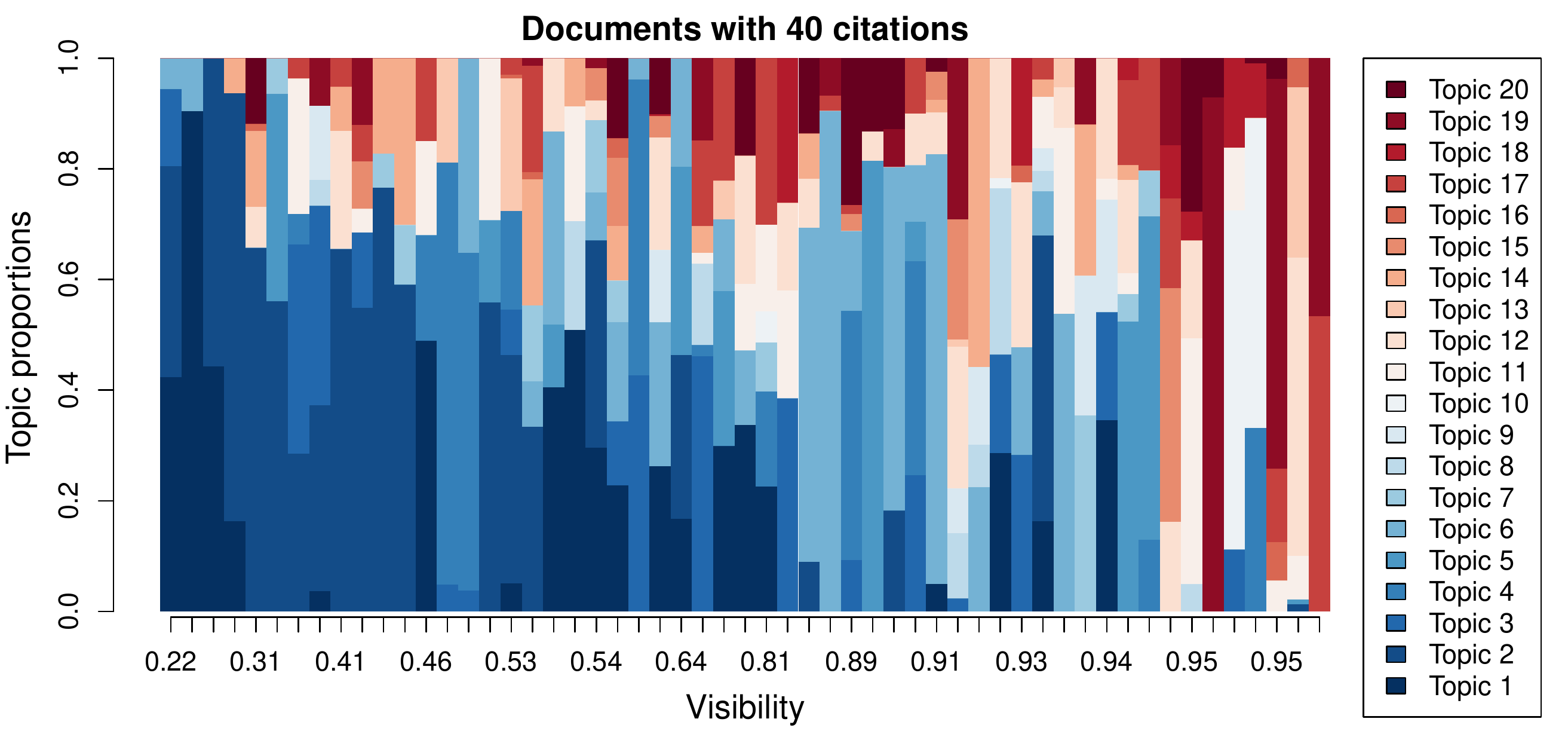}
\caption {\label{KDDlqbp} KDD barplot: each bar represent the topic proportions of a document with 40 citations. The documents are arranged by visibility in increasing order. The legend indicates the color representing each topic. The topics are ordered by average number of citations received; topic 1 has the highest and topic 20 the lowest average number of citations. This figure is intended to be read in color (color version available in online publication).}
\end{figure}
Figure \ref{KDDlqbp} is a barplot showing the topic proportions of documents with 40 citations. We have fixed the number of citations in order to study how visibility varies with topic proportions. As visibility increases, there is a transition in the color of the bars from being dominantly blue to red. That is, for the {\em same citation count} (40 in this case), the estimated visibility of an article from topics with low citation rates (red) tends to be greater than an article from topics with high citation rates (blue). This phenomenon demonstrates that the visibility metric adjusts for differences in citation frequency among different topics by assigning higher visibilities to articles from topics with lower citation rates for the same citation count. For example, consider an article from Mathematics and another from Molecular Biology having the same number of citations. As the citation rate in Mathematics is much lower than in Molecular Biology, it is inappropriate to compare the raw citation counts of these articles directly. Citations in Mathematics ought to be assigned higher weights and this can be achieved through normalization procedures, for example by dividing citation counts by the average number of citations per article for a discipline \citep{Radicchi2008}. However, the choice of a suitable reference standard is a very intricate issue. The topic-adjusted visibility metric that we propose offers an alternative to tackle the problem of field variation.

\section{Conclusion} \label{conclusion}
Citation activities of scientific applications and other article-centric activities (shares on social web, for instance) are of interest for the evaluation of scientific merit and impact of published research. In particular, the citation network among published articles is a special case of a relational network. Communities detected in a citation network has been shown to correlate well with scientific areas and research topics. A unique feature of a citation network is that content information is available on individual nodes, which is arguably influenced by the same mixed group structure underpinning the citation connectivity. Combining content and connectivity information in a citation network may alleviate the multi-modality issue of community detection in network analysis. On the other hand, communities detected in the citation network regularize the topic modeling on the content information. 

The probability of being cited for a particular article is determined not only by its membership in one or more topic domains and the topic-level citation rates (within and between domains), but also factors that are unique to this publication such as timing, visibility of the authors, and novelty and importance of the results. In this paper, we introduce a model for citation networks that infers the topic domain structure of the articles and their citation links, and estimates the citation activity rates at both the topic domain level and the article level. For each article, we introduce a latent variable that serves as a topic-adjusted visibility metric of this article. A higher value of this latent metric indicates that this article is more likely to be cited than other articles that are located close-by in the topic domain.  As we have shown in our application to real datasets, this metric correlates with raw citations counts but is not merely a measure of popularity, as it accounts for variation in activity levels among different topics. The proposed model leads to significant improvement in link predictions, which can be helpful in article recommendation. It provides a better visibility metric for comparing individual scientific publications across different fields.  

The inference of the proposed model is realized via a novel variational Bayes algorithm. For real-world large document networks, we propose a sub-sampling strategy that enables the use of stochastic variational inference, which is computationally efficient and achieves a similar level of predictive performance as the variational Bayes algorithm.

\section*{Acknowledgments}
Tian Zheng's research is supported by NSF grant SES-1023176. Linda Tan's research is supported by the National University of Singapore Overseas Postdoctoral Fellowship. We thank the editor, associate editor and the reviewers for their comments which have improved the manuscript greatly.

\begin{supplement}[id=suppA]
  \sname{Supplement}
  \stitle{“Topic-adjusted visibility metric for scientific articles”}
  \slink[doi]{COMPLETED BY THE TYPESETTER}
  \sdatatype{.pdf}
  \sdescription{We provide additional material to support the results in this paper. This include further discussions, detailed derivations, illustrations and a simulation study.}
\end{supplement}


\newpage

\begin{frontmatter}
	
	\title{Supplement to ``Topic-adjusted visibility metric for scientific articles"}
	\runtitle{Supplement}
	\begin{aug}
		\author{\fnms{Linda S. L.}  \snm{Tan}\thanksref{m1,m2}\ead[label=e1]{st2924@columbia.edu}},
		\author{\fnms{Aik Hui} \snm{Chan}\thanksref{m2}\ead[label=e2]{phycahp@nus.edu.sg}}
		\and
		\author{\fnms{Tian}  \snm{Zheng}\thanksref{t1,m1}
			\ead[label=e3]{tzheng@stat.columbia.edu}%
			\ead[label=u1,url]{http://www.foo.com}}
		
		\thankstext{t1}{Corresponding author: tzheng@stat.columbia.edu}
		\runauthor{Tan, Chan and Zheng}
		\affiliation{Columbia University\thanksmark{m1} and National University of Singapore\thanksmark{m2}}
		
	\end{aug}
	
\end{frontmatter}

\setcounter{section}{0} \renewcommand{\thesection}{S\arabic{section}}
\setcounter{figure}{0} \renewcommand{\thefigure}{S\arabic{figure}}
\setcounter{table}{0} \renewcommand{\thetable}{S\arabic{table}}

\section{Variational inference for the LMV}
In the variational approximation for the LMV, we assume that (1) the variational posterior is of a factorized form, and (2) variational posteriors of $B_{ij}$ and $\tau_{d'}$ are in the same family as their priors. The other distributions arise naturally as the optimal densities under the product density restriction. For $B_{ij}$ and $\tau_{d'}$, the likelihood is nonconjugate with respect to the prior. Hence we have assumed the above parametric distributions. The factorization assumption may have a degrading effect on inference if the variables have a high degree of dependence in the posterior. However, if posterior dependence between variables is weak, then variational approximation can lead to very accurate approximate inference. It has been observed that variational approximation is often able to capture the means accurately but tends to underestimate the posterior variance \citep{Wang2005, Bishop2006}. Variational approximation can also lead to excellent predictive inference \citep[e.g. large-scale discrete choice models,][]{Braun2010}. Note that in variational inference, the fitted model is only guaranteed to be optimal locally (not necessarily globally). A common strategy to overcome this issue is to run the variational algorithm starting with random initialization from multiple starting points. 

\subsection{First-order approximation about the mean}
In the variational inference for the LMV, we encounter the term $E_q\{ \log (1-\tau_{d'} B_{ij}) \}$ which cannot be evaluated analytically, and we approximate it by taking the expectation of a first-order Taylor series expansion about the mean. Let $f(X)=\log(1-X)$, $X=\tau_{d'} B_{ij}$ and $\mu=\text{E}_q(X)$, where $\text{E}_q$ denotes expectation with respect to the variational posterior. It is assumed in $q$ that $\tau_{d'}$ and $B_{ij}$ are independent, $q(\tau_{d'})=\text{Beta}(g_{d'}, h_{d'})$ and $q(B_{ij})=\text{Beta}(a_{ij}, b_{ij})$. For a Beta$(\alpha, \beta)$ random variable $Y$, $E(Y^k)=\frac{\alpha+k-1}{\alpha+\beta+k-1}E(Y^{k-1}) < 1$. Therefore, the Taylor series of $X$ is well-defined as all the moments of $X$ are bounded and decreases as $k$ increases. Taking expectation of the Taylor series of $f(X)$ about $\mu$, we have
\begin{equation} \label{approx}
\begin{aligned}
f(X) &= f(\mu) +f'(\mu)(X-\mu)+\frac{f''(\mu)}{2!}(X-\mu)^2+\dots, \\
\Rightarrow \text{E}_q\{f(X)\} &= f(\mu) +\frac{f''(\mu)}{2}\text{Var}_q(X)+\dots.
\end{aligned}
\end{equation}
The first-order approximation $\text{E}_q\{f(X)\} \approx  f(\mu)$ is exact when the support of $X$ converges to a point. From Jensen's inequality, we also have $\text{E}_q\{f(X)\} \leq  f(\mu)$ as $f$ is concave. While the first-order approximation is crude, we reason below that it often works well in our context.

The $k$th term in the second line of \eqref{approx} is given by $-\text{E}_q \{(X-\mu)^k\} / k(1-\mu)^k$. In the limit as $\mu \rightarrow 0$, this term approaches $-\text{E}_q (X^k) / k$ and thus the error in the first-order approximation is bounded by $\mathcal{O}(\text{E}_q(X^2))$. Note that $\mu \rightarrow 0$ implies $h_{d'} \gg g_{d'}$ or $b_{ij} \gg a_{ij}$, in which case, $\text{E}_q(X^2) \rightarrow 0$ as 
\begin{equation*}
\text{E}_q(X^2) = \text{E}_q(\tau_{d'}^2)  \text{E}_q(B_{ij}^2) = \frac{g_{d'}+1}{g_{d'}+h_{d'}+1} \frac{g_{d'}}{g_{d'}+h_{d'}} \frac{a_{ij}}{a_{ij}+b_{ij}} \frac{a_{ij}+1}{a_{ij}+b_{ij}+1}.
\end{equation*}
In the context of document networks, it is often the case that $\mu$ is close to 0 as the sparsity of the network (small number of observed links) translates into a low probability of a document from topic $i$ citing another document from topic $j$. Since this probability is approximately $a_{ij}/(a_{ij}+b_{ij})$ in the variational posterior, $b_{ij}$ is usually much larger than $a_{ij}$.

\subsection{Approximate lower bound} \label{appendix - Lower bound}
We present the complete expression for the approximate lower bound  $\mathcal{L}^*$ discussed in Section \ref{posterior inference}. Using \eqref{lower bound} and \eqref{first-order approx}, $\mathcal{L}^*$ can be evaluated as
\begin{equation*}
\begin{aligned}
&\sum_{k,v} \bigg(\negmedspace\eta_v + \sum_{d,n} w_{dnv} \phi_{dnk} -\lambda_{kv} \bigg) \bigg[\psi(\lambda_{kv})-\psi\bigg(\sum _v \negmedspace\lambda_{kv}\negmedspace\bigg)\bigg] - \sum_d \log \Gamma\bigg(\negmedspace\sum_k \gamma_{dk}\negmedspace\bigg)   \\
& - \negmedspace\sum_{d, n, k} \phi_{dnk} \log \phi_{dnk} + \sum_{i,j}  \{\log B(a_{ij}, b_{ij}) -  \log B(a_0, b_0) \} + \sum_{d,k}  \log \Gamma(\gamma_{dk}) \\
& + \sum_{d,k} \bigg(\alpha_k+ \sum_n \phi_{dnk}+  \sum_{d' \neq d} (\kappa_{dd'k} + \nu_{d'dk})  -\gamma_{dk} \bigg) \bigg[\psi(\gamma_{dk})-\psi\bigg(\sum_{k} \gamma_{dk}\bigg)\bigg] \\
& + D \bigg[  \log \Gamma\bigg(\negmedspace\sum_{k} \alpha_k\negmedspace\bigg) - \sum_{k} \log \Gamma(\alpha_k) \bigg] + K \bigg[  \log \Gamma\bigg(\sum_v \eta_v \bigg) - \sum_v \log \Gamma(\eta_v) \bigg]\\
& + \sum_{(d,d')} \sum_{i,j} \kappa_{dd'i} \nu_{dd'j} \Big\{ y_{dd'} \left[ \psi(a_{ij}) - \psi(a_{ij} + b_{ij}) + \psi(g_{d'}) - \psi(g_{d'} + h_{d'}) \right] \\
& +  (1-y_{dd'}) \log \left(1- \tfrac{g_{d'}}{g_{d'}+h_{d'}} \tfrac{a_{ij}}{a_{ij} +b_{ij}} \right)\negmedspace \Big\} + \sum_{k, v}  \log \Gamma(\lambda_{kv}) -  \sum_k \log \Gamma\bigg(\negmedspace\sum_v \lambda_{kv} \negmedspace\bigg)  \\
&+ \sum_{i,j}\{(a_0 - a_{ij})[\psi(a_{ij}) - \psi(a_{ij} + b_{ij})] \negmedspace +\negmedspace ( b_0 - b_{ij})[\psi(b_{ij}) - \psi(a_{ij} + b_{ij}) ]\}  \\
& + \sum_{d'}\{ (g_0 - g_{d'}) [\psi(g_{d'}) - \psi(g_{d'} + h_{d'}) ] + (h_0 - h_{d'})  [\psi(h_{d'}) - \psi(g_{d'} + h_{d'}) ]\} \\
& + \sum_{d'} \log B (g_{d'}, h_{d'}) - \negmedspace D\negthinspace \log B(g_0, h_0) -\negthickspace \sum_{(d,d'), k} \negthickspace(\kappa_{dd'k} \log \kappa_{dd'k} + \nu_{dd'k} \log \nu_{dd'k}),
\end{aligned}
\end{equation*}
where $\Gamma(\cdot)$, $\psi(\cdot)$ and $B(\cdot, \cdot)$ denote the gamma, digamma and beta functions respectively.

\subsection{Deriving updates} \label{appendix - NCVMP updates}
Next, we show how closed form updates for $\{a, b, g, h\}$ discussed in Section \ref{posterior inference} can be derived using nonconjugate variational message passing. First, we review some general results. Suppose $q(\Theta)=\prod_{i=1}^m q_i(\Theta_i)$ and $q_i(\Theta_i)$ is a member of some exponential family such that
\begin{equation*}
q_i(\Theta_i)=\exp\{\lambda_i^T t_i(\Theta_i)-h_i(\lambda_i)\},
\end{equation*}
where $\lambda_i$ is the vector of natural parameters and $t_i(\cdot)$ are the sufficient statistics. The ordinary gradient of the lower bound $\mathcal{L} = E_q \{ \log p(y, \Theta)  \} - E_q \{ \log q(\Theta) \}$ with respect to $\lambda_i$ is 
\begin{equation*}
\nabla_{\lambda_i}\mathcal{L} =  \nabla_{\lambda_i} E_q\{\log p(y,\Theta)\} - \text{Cov}_{q_i} [t_i(\Theta_i)] \lambda_i,
\end{equation*}
where $\text{Cov}_{q_i} [t_i(\Theta_i)] $ represents the covariance matrix of $t_i(\Theta_i)$. As $q_i(\Theta_i)$ is an exponential family member, $\text{Cov}_{q_i} [t_i(\Theta_i)] $ is equal to the Fisher information matrix of $q_i$. The natural gradient can be computed by pre-multiplying the ordinary gradient with the inverse of the Fisher information matrix \citep{Amari1998}. Thus the natural gradient of $\mathcal{L}$ with respect to $\lambda_i$ is given by
\begin{equation} \label{nat_grad}
\tilde{\nabla}_{\lambda_i} \mathcal{L} = \text{Cov}_{q_i}\{t_i(\Theta_i)\}^{-1} \nabla_{\lambda_i} E_q \{ \log p(y, \Theta) \} - \lambda_i
\end{equation}
Setting $\tilde{\nabla}_{\lambda_i} \mathcal{L}=0$ leads to the fixed point update:
\begin{equation} \label{NCVMP_update}
\lambda_i \leftarrow \text{Cov}_{q_i} [t_i(\Theta_i)] ^{-1} \nabla_{\lambda_i} E_q\{\log p(y,\Theta)\}.
\end{equation}
In Section \ref{stochastic_gradient_updates}, we write the natural gradient in \eqref{nat_grad} as $\hat{\lambda}_i - \lambda_i$, where $\hat{\lambda}_i = \text{Cov}_{q_i} [t_i(\Theta_i)] ^{-1} \nabla_{\lambda_i} E_q\{\log p(y,\Theta)\}$ is simply the nonconjugate variational message passing update in \eqref{NCVMP_update}. In the case of conjugate priors, $\hat{\lambda}_i $ is independent of $\lambda_i$ and reduces to the update of $\lambda_i$ in variational Bayes. See for example \cite{Tan2014}. 

The variational posteriors $q(B_{ij})=\text{Beta}(a_{ij}, b_{ij})$ and $q(\tau_{d'}) = \text{Beta}(g_{d'}, h_{d'})$ belong to the exponential family. Thus we can derive updates for them using \eqref{NCVMP_update}. The natural parameter of $\text{Beta}(a,b)$ is $[a-1 \;\;  b-1]^T$. Some useful results which can be easily verified are stated below.
\begin{results}
	Let $I_{a,b} = \begin{bmatrix} \psi'(a) - \psi' (a+b) & -\psi'(a+b) \\ -\psi'(a+b) & \psi'(b) - \psi' (a+b) \end{bmatrix}$ be the \\
	Fisher information matrix of $\text{Beta}(a,b)$.
	\begin{enumerate}[label=(\alph*)]
		\item $I_{a,b} ^{-1} \begin{bmatrix} (a_0-1) [\psi'(a) - \psi'(a+b)] - (b_0-1) \psi'(a+b) \\ -(a_0-1) \psi'(a+b) + (b_0-1) [\psi'(b) - \psi'(a+b)] \end{bmatrix} = \begin{bmatrix} a_0 -1 \\ b_0 -1 \end{bmatrix}$
		\item $I_{a,b} ^{-1} \begin{bmatrix} \psi'(a) - \psi'(a+b) \\ - \psi'(a+b) \end{bmatrix} = \begin{bmatrix} 1 \\ 0 \end{bmatrix}$
	\end{enumerate}
\end{results}

Using \eqref{NCVMP_update} and Results 1, the updates of $a_{ij}$ and $b_{ij}$ are given by
\begin{equation*}
\begin{aligned}
&\begin{bmatrix} \hat{a}_{ij} -1 \\ \hat{b}_{ij} -1 \end{bmatrix} = I_{a_{ij},b_{ij}} ^{-1} \begin{bmatrix} \nabla_{a_{ij}} E_q \{\log p(y, \Theta)\} \\ \nabla_{b_{ij}} E_q \{\log p(y, \Theta)\}  \end{bmatrix} \\
& =  I_{a_{ij},b_{ij}} ^{-1} \left\{  \begin{bmatrix}
(a_0-1)\{\psi'(a_{ij}) - \psi'(a_{ij} + b_{ij}) \} -  (b_0-1)  \psi'(a_{ij} + b_{ij}) \\ -  (a_0-1) \psi'(a_{ij} + b_{ij}) + (b_0-1) \{\psi'(b_{ij}) - \psi'(a_{ij} + b_{ij})\} 
\end{bmatrix} \right.\\
& \quad  \left. + \begin{bmatrix} \psi'(a_{ij}) - \psi'(a_{ij} + b_{ij})  \\  -\psi'(a_{ij} + b_{ij})  \end{bmatrix} 
\sum_{\substack{(d,d'):\\ y_{dd'}=1}} \negthickspace \kappa_{dd'i}\nu_{dd'j}  + \begin{bmatrix} -b_{ij} \\ a_{ij}  \end{bmatrix}  \sum_{\substack{(d,d'):\\ y_{dd'}=0}} \negthickspace \frac{ \frac{ \kappa_{dd'i}\nu_{dd'j} }{(a_{ij} +b_{ij})^2} \frac{g_{d'}}{g_{d'}+h_{d'}}  }{1- \frac{g_{d'}}{g_{d'}+h_{d'}} \frac{a_{ij}}{a_{ij} +b_{ij}} }  \right\}\\
& = \begin{bmatrix} a_0 -1 \\ b_0 -1 \end{bmatrix} + \begin{bmatrix} \sum_{(d,d') \in S_1} \kappa_{dd'i}\nu_{dd'j}  \\ 0 \end{bmatrix} 
+ \frac{1}{(a_{ij} +b_{ij})^2 |I_{a_{ij},b_{ij}}|}\;   \\
& \quad \times \begin{bmatrix} (a_{ij} + b_{ij}) \psi'(a_{ij} + b_{ij}) - b_{ij} \psi'(b_{ij}) \\ a_{ij} \psi'(a_{ij}) -(a_{ij} + b_{ij}) \psi'(a_{ij} + b_{ij}) \end{bmatrix}  
\sum_{d'=1}^D  \frac{ \{\sum_{d: y_{dd'}=0} \kappa_{dd'i}\nu_{dd'j}\} \frac{g_{d'}}{g_{d'}+h_{d'}}  }{1- \frac{g_{d'}}{g_{d'}+h_{d'}} \frac{a_{ij}}{a_{ij} +b_{ij}} }.
\end{aligned}
\end{equation*}
where $|I_{a,b}| = \psi'(a)\psi'(b) - \psi'(a + b\{ \psi'(a) + \psi'(b)  \}$.

Similarly, 
\begin{equation*}
\begin{aligned}
&\begin{bmatrix} \hat{g}_{d'} -1 \\ \hat{h}_{d'} -1 \end{bmatrix} = I_{g_{d'},h_{d'}} ^{-1} \begin{bmatrix} \frac{\partial E_q \{\log p(y, \Theta)\} }{\partial g_{d'}} \\ \frac{\partial E_q \{\log p(y, \Theta)\} }{\partial h_{d'}} \end{bmatrix} \\
& =  I_{g_{d'},h_{d'}} ^{-1}  \left\{  \begin{bmatrix}
(g_0-1)\{\psi'(g_{d'}) - \psi'(g_{d'} + h_{d'}) \} -  (h_0-1)  \psi'(g_{d'} + h_{d'}) \\ -  (g_0-1) \psi'(g_{d'} + h_{d'}) + (h_0-1) \{\psi'(h_{d'}) - \psi'(g_{d'} + h_{d'})\} 
\end{bmatrix} \right. \\
&  \quad \left.+  \begin{bmatrix} \psi'(g_{d'}) - \psi'(g_{d'} + h_{d'})  \\  -\psi'(g_{d'}+ h_{d'})  \end{bmatrix} \negthickspace \sum_{d: y_{dd'}=1} \negmedspace y_{dd'} +  \begin{bmatrix}
-h_{d'} \\ g_{d'}  \end{bmatrix} \sum_{i,j}   \frac{ \sum_{d: y_{dd'}=0}  \frac{\kappa_{dd'i}\nu_{dd'j}}{(g_{d'} +h_{d'})^2} \frac{a_{ij}}{a_{ij} +b_{ij}} }{1- \frac{g_{d'}}{g_{d'}+h_{d'}}  \frac{a_{ij}}{a_{ij} +b_{ij}} } \right\}\\
& = \begin{bmatrix} g_0 -1 \\ h_0 -1 \end{bmatrix} + \begin{bmatrix} \sum_{d: y_{dd'}=1}  y_{dd'}  \\ 0 \end{bmatrix} 
+ \frac{1}{(g_{d'} +h_{d'})^2 |I_{g_{d'},h_{d'}}|}\;   \\
& \quad \times \begin{bmatrix} (g_{d'} + h_{d'}) \psi'(g_{d'} + h_{d'}) - h_{d'} \psi'(h_{d'}) \\ g_{d'} \psi'(g_{d'}) -(g_{d'} + h_{d'}) \psi'(g_{d'} + h_{d'}) \end{bmatrix}  
\sum_{i,j}   \frac{\{ \sum_{d: y_{dd'}=0} \kappa_{dd'i}\nu_{dd'j}\} \frac{a_{ij}}{a_{ij} +b_{ij}} }{1- \frac{g_{d'}}{g_{d'}+h_{d'}}  \frac{a_{ij}}{a_{ij} +b_{ij}} } .
\end{aligned}
\end{equation*}

\section{Illustration of sampling strategy} \label{sampling_strategy_illustration}
\begin{figure}[b]
	\centering
	\includegraphics[width=0.55\textwidth]{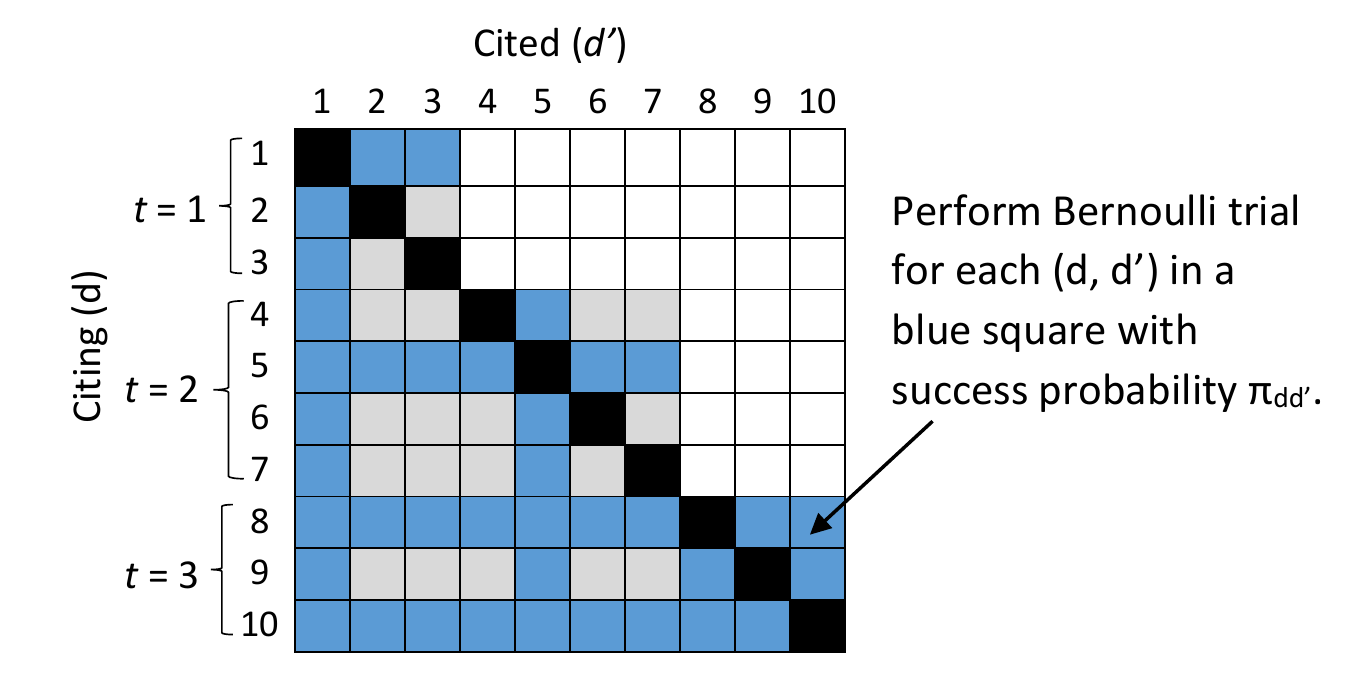}
	\caption {\label{sample} Documents 1--10 published in three time periods, $t=1,2,3$. Suppose $S=\{1, 5, 8, 10\}$. Grey squares represent document pairs $(d,d')$ for which links might be observed. Bernoulli trials are performed for document pairs in blue squares. Successful pairs are included in $\mathcal{P}$.}
\end{figure}

We use a simple example to illustrate how the sampling strategy described in Section \ref{sampling_strategy} can be implemented in the case where publication times are available. Suppose we have a dataset with $D=10$ documents published in three time periods, say documents 1--3, 4--7 and 8--10 are published respectively in the periods $t=1$, 2 and 3. Assuming that documents can only cite other documents published in the same or earlier time periods, the block lower triangular matrix in Figure \ref{sample} represent all document pairs where links might be observed. Suppose a random sample $S=\{1, 5, 8, 10\}$ is drawn, Bernoulli trials are performed for document pairs in the blue squares and successful pairs are included in $\mathcal{P}$. 

\section{Horvitz-Thompson estimators}
The estimators in \eqref{HT estimators} are constructed through the use of weighted averaging. To show that 
\begin{equation*}
\text{E} \Bigg( \frac{D}{|\mathcal{S}|}\negthinspace \sum_{\substack{(l,l') \in \mathcal{P}_{\Cdot d'} \\ d' \in \mathcal{S}} } \negthickspace \negthickspace \negthickspace \frac{\kappa_{ll'i}\nu_{ll'j}}{\pi_{ll'}} \Bigg) = \sum_{(d,d')} \kappa_{dd'i}\nu_{dd'j}, 
\end{equation*}
imagine that document pairs are selected through the following two-stage cluster sampling procedure P: 
\begin{enumerate}[leftmargin=1.3em]
	\item Randomly select a sample $\mathcal{S}$ of documents. The columns corresponding to these documents in the adjacency matrix are the clusters selected at the first stage.
	\item Perform a Bernoulli trial with success probability $\pi_{ll'}$ for each document pair $(l,l')$ in these columns (i.e. for $l' \in \mathcal{S}$) and select those pairs that are successful. 
\end{enumerate}
Now, introduce a binary random variable $Z_{(l,l')}$ for each document pair $(l,l')$ such that $Z_{(l,l')}=1$ if $ (l,l')$ is selected through procedure P and 0 otherwise. Then\begin{equation*}
P(Z_{(l,l')}=1) =P(d' \in \mathcal{S})P((l,l') \in \mathcal{P}_{\Cdot d'}|d' \in \mathcal{S}) =\frac{|\mathcal{S}|}{D} \pi_{ll'}.
\end{equation*}
Therefore
\begin{multline*}
\text{E} \Bigg( \frac{D}{|\mathcal{S}|}\negthinspace \sum_{\substack{(l,l') \in \mathcal{P}_{\Cdot d'} \\ d' \in \mathcal{S}} } \negthickspace \negthickspace \negthickspace \frac{\kappa_{ll'i}\nu_{ll'j}}{\pi_{ll'}} \Bigg) = \text{E} \bigg( \frac{D}{|\mathcal{S}|} \sum_{(l,l')} \frac{\kappa_{ll'i}\nu_{ll'j}}{\pi_{ll'}} Z_{ll'}  \bigg) \\
=  \frac{D}{|\mathcal{S}|} \sum_{(l,l')} \frac{\kappa_{ll'i}\nu_{ll'j}}{\pi_{ll'}} E(Z_{ll'})  = \sum_{(d,d')} \kappa_{dd'i}\nu_{dd'j}.
\end{multline*}

\section{Implementing the stochastic variational algorithm}
At every iteration of the stochastic variational algorithm, we randomly sample a minibatch $\mathcal{S}$ of documents and update document specific variational parameters only for $d \in \mathcal{S}$. Sampling with or without replacement may be used as both approaches produce unbiased estimates. To ensure that document specific variational parameters are updated uniformly, we sample randomly without replacement at each iteration.
That is, starting with $D$ documents, we randomly sample $|\mathcal{S}|$ documents in the first iteration. At the second iteration, we sample another $|\mathcal{S}|$ documents from the remaining $D- |\mathcal{S}|$ documents. This continues for say $M$ iterations until there are no documents left and we consider this as one sweep through the corpus. All documents are then replaced and we start making a second sweep. This process is repeated until convergence. In this approach, all document specific variational parameters are updated once in each sweep.

Recall that the updates of $\{a, b\}$ and $\{g,h\}$ are highly interdependent and hence a nested loop was introduced in Algorithm 1 for these parameters. While it is not possible to use a nested loop in Algorithm 2 (conditions for convergence of stochastic gradient algorithms will not be satisfied), we find that using moderately smaller minibatch sizes alleviates this issue as the frequency of feedback between these updates is increased within each sweep. 

We use a step size of the form $\frac{A_1}{(s_w+\frac{m}{M}+A_2)^v}$, where $M$ denotes the number of iterations required to make a sweep through the corpus and $0 \leq m \leq M-1$ denotes the number of minibatches that have been analyzed at the $s_w$th sweep. This step size is of the form $s_t=\frac{w}{(t+W)^v}$ with $w=A_1M^v$ and $W= M(A_2+1)-1$. For the Cora and KDD datasets, we set $A_1=1$, $A_2=5$, $v= 0.501$ for the stepsize.

\section{Priors and stopping criteria}
The same priors are used across all models. All hyperparameters of beta distributions are set as 1. We have tried more informative priors for the blockmodel but observed that the algorithms are not very sensitive to the hyperparameter values. In the real examples, we set $\alpha=[1/K, \dots, 1/K]$. A symmetric Dirichlet($\eta$) prior is used for each $\beta_k$. In our applications, we observed that small $\eta$ leads to better predictive performance for the RTM while a larger $\eta$ works better for Pairwise-Link-LDA and the LMV. This is likely due to the difference in modeling assumptions; RTM considers a diagonal weight matrix while Pairwise-Link-LDA and LMV use a blockmodel. 

All algorithms running in batch mode are stopped when the relative increase in the lower bound is less than $10^{-5}$. For stochastic algorithms, we consider the relative change in the diagonal of the blockmodel after each sweep and stop when $\| \text{diag}(B^{(s_w)} - B^{(s_w -1) }) \| < \varepsilon \| \text{diag}(B^{(s_w -1) }) \|$, where $\| \cdot \|$ denotes the Euclidean norm and $\varepsilon$ is some small tolerance. We set $\varepsilon=0.05$ for the Cora dataset and $\varepsilon=0.1$ for the KDD dataset.

\section{Examining quality of latent topics}
The quality of uncovered latent topics may be validated by examining the coherence of top words in each topic, comparing topics or identifying documents that are most representative of each topic. \cite{Chuang2012} present a visualization system called Termite that supports evaluation of topics by introducing measures for ranking and sorting terms. \cite{Wallach2009} propose and compare methods for evaluating topic models based on estimating the probability of held-out documents given a trained model, while \cite{Mimno2011} present a new Bayesian method based on posterior predictive checking for measuring how well a topic model fits a corpus.

\section{Simulation study} \label{sim_eg}
\begin{figure}
	\centering
	\includegraphics[width=0.9\textwidth]{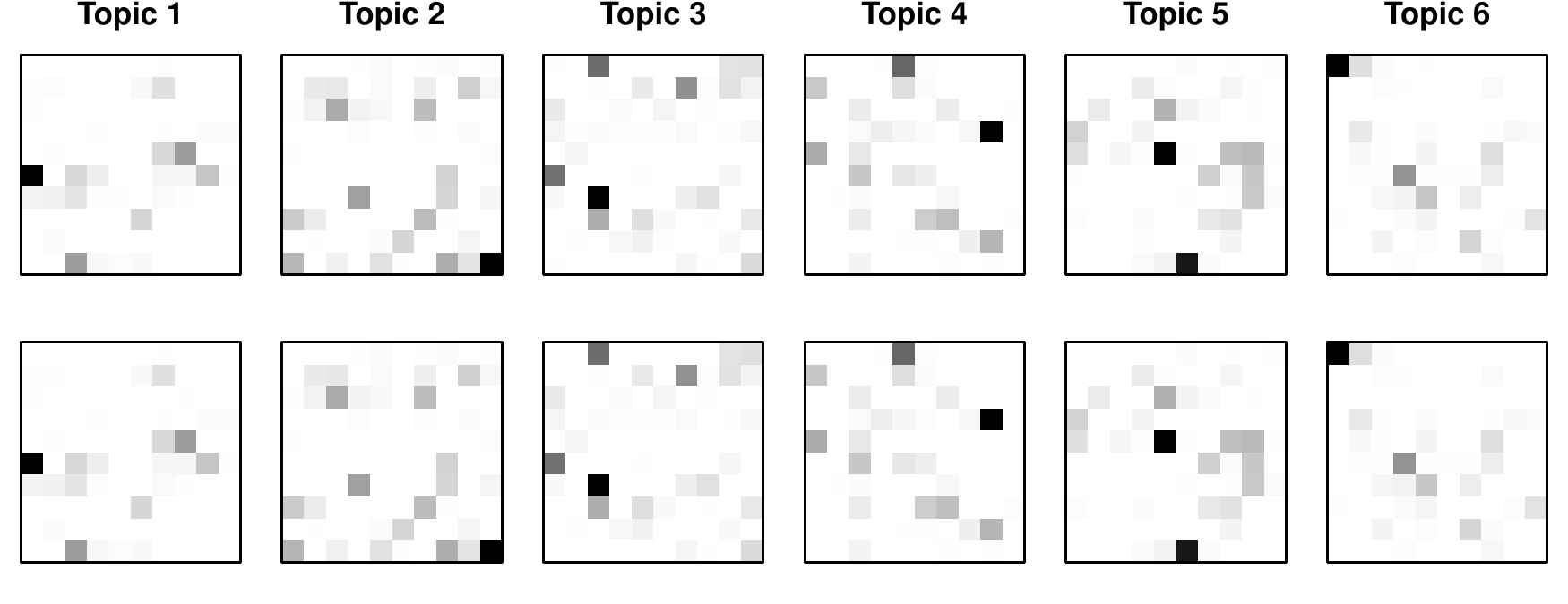}
	\caption {\label{simtopics} Graphical representation of $K=6$ topics. First row: simulated topics. Second row: topics found by LMV (Algorithm 1).}
\end{figure}

We present here results of the simulation study mentioned in Section \ref{examples}. To evaluate the accuracy and computational performance of Algorithms 1 and 2, we generate 20 datasets each consisting of $3000$ documents from the LMV. We consider $K=6$ topics generated from the symmetric Dirichlet(0.1) distribution. These topics are represented using images in Figure \ref{simtopics} (first row). Each image contains $V=100$ pixels in a $10 \times 10$ grid and each pixel corresponds to a term in the vocabulary. The intensity of a pixel represents its frequency (grid is for ease of reading). We note that the text may not have been simulated in a realistic manner. However, we wanted an example where the topics can be visualized and recovered easily so that we can test the abilities of Algorithms 1 and 2 in recovering the blockmodel and visibilities. When the simulated topics are not sufficiently distinct or if there are overlaps, the inferred topics may turn out to be some combination of the simulated topics. Multiple runs may also produce varying results. In such cases, it will then not be possible to recover the true blockmodel and visibilities. To avoid such scenarios, we have used an example with a small number of well differentiated topics so that we can evaluate the accuracy of the recovered blockmodel and visibilities and study the behaviour of different algorithms on a common basis.

\begin{figure}[tb!]
	\centering
	\begin{minipage}[c]{0.58\textwidth}
		\centering
		\begin{tabular}{|rrrrrr|} 
			\hline
			\textbf{0.3}   & 0     & 0.02 & 0     & 0       & 0.03  \\
			0      & \textbf{0.3}  & 0      & 0      & 0.05 & 0       \\
			0      &  0    & \textbf{0.3}   & 0.05 & 0      & 0        \\
			0.04 & 0     &  0     &\textbf{ 0.2}   & 0      & 0.02   \\
			0       & 0.04 & 0     & 0      & \textbf{0.2}  & 0        \\
			0.02  & 0     & 0.03 & 0      & 0      & \textbf{0.2}   \\ 
			\hline
		\end{tabular}
	\end{minipage}
	\begin{minipage}[c]{0.3\textwidth}
		\centering
		\includegraphics[width=0.8\textwidth]{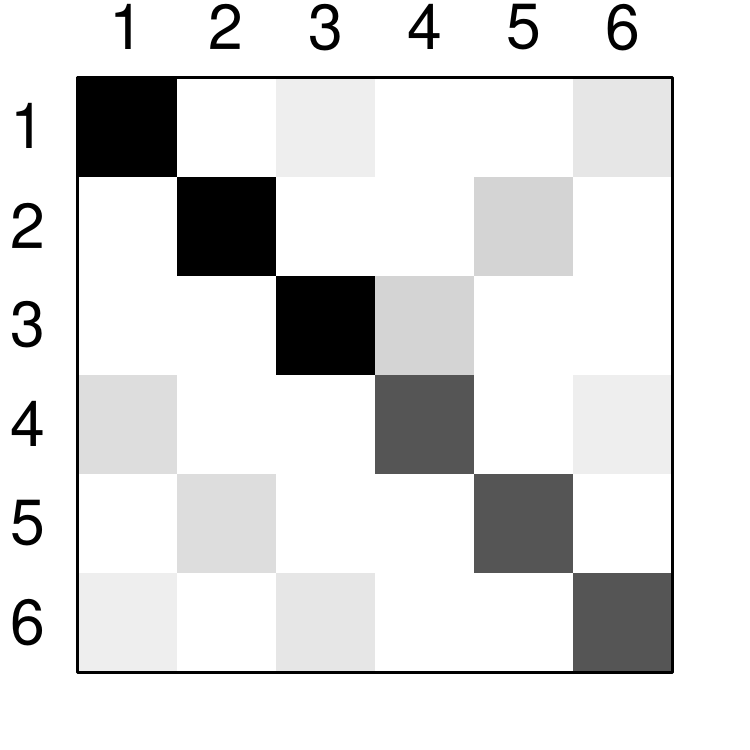}
	\end{minipage}
	\caption{True blockmodel $B$.}
	\label{blockmodel}
\end{figure}

Topic assignments are generated from symmetric Dirichlet(0.05) and the visibility of each document is generated from Beta(1, 1). We consider the blockmodel $B$ shown in Figure \ref{blockmodel}. The table on the left shows the values in $B$ while the image on the right is a visualization of $B$. The values on the diagonal are larger, indicating a higher probability of within-topic citations. There are also certain off-diagonal elements which are nonzero but smaller in value indicating a smaller probability of citations across topics. 
Each of the 20 datasets differ in terms of links and text but are simulated using the same topics, blockmodel and visibilities. The number of words in each document is 100. The first 2000 documents are used for training and the remaining 1000 documents are used as the test set. 

\begin{figure}[tb!]
	\centering
	\includegraphics[width=0.7\textwidth]{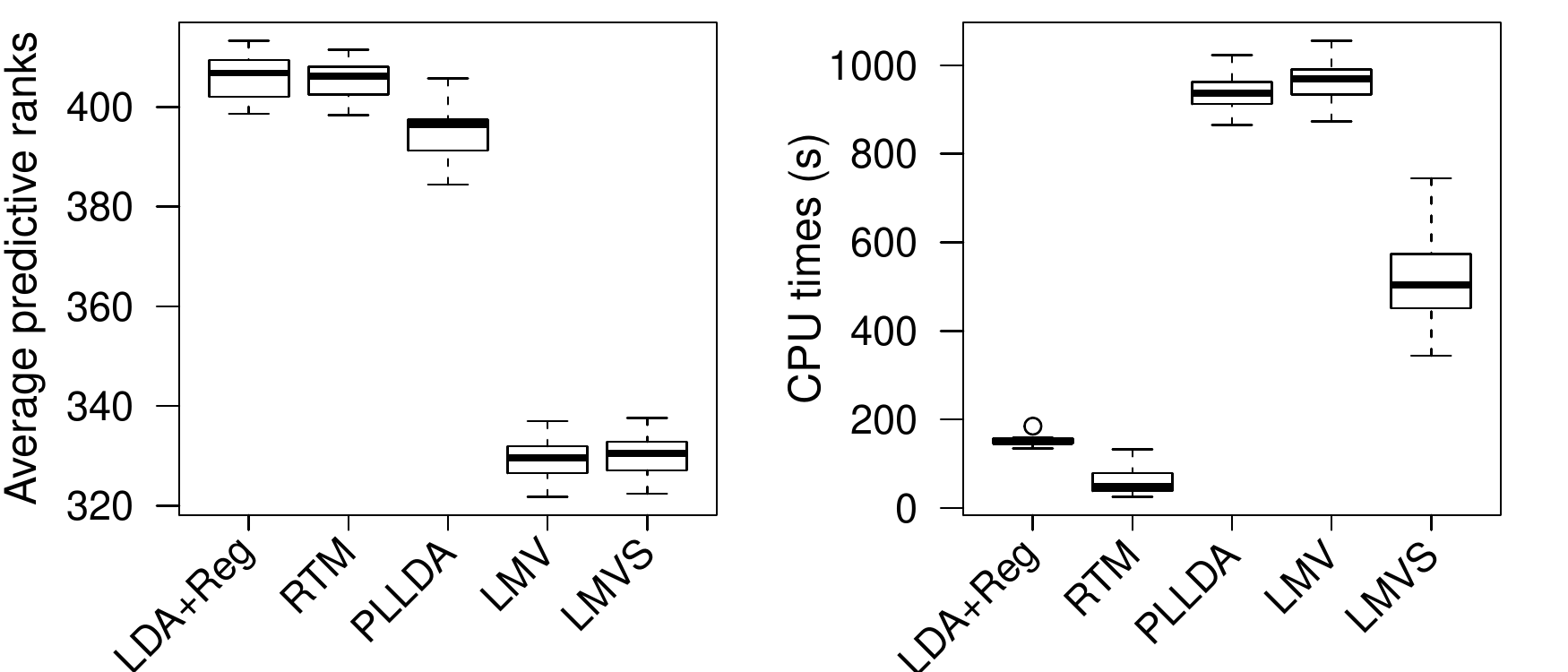}
	\caption {\label{simranktime} Barplots of average predictive ranks (left) and CPU times (right) for 20 simulated datasets. }
\end{figure}

The simulated datasets are fitted using LDA+Regression, RTM, Pairwise-Link-LDA and Algorithms 1 and 2. For Algorithm 2, we let the mini-batch size be 200 and $\varepsilon=0.015$. For the step size, we let $A_1=2$, $A_2=3$ and $v=0.501$. Hyperparameters used to generate the datasets were used as prior hyperparameters. All methods were able to recover topics that were close to the ones used to simulate the data. The topics found by LMV (Algorithm 1) are shown in the second row of Figure \ref{simtopics}. The average predictive ranks and CPU times of different methods are shown in Figure \ref{simranktime}. 
Average predictive ranks of LDA+Regression and RTM are similar while Pairwise-Link-LDA does slightly better. LMV yields the lowest average predictive ranks and results produced by Algorithms 1 and 2 are very close. In this example, the true blockmodel $B$ has some nonzero off-diagonal elements which can be captured by Pairwise-Link-LDA and the LMV, but LDA+Regression and the RTM cannot model citations across different topics. LMV also performed better than Pairwise-Link-LDA because the simulated data use visibilities to distinguish the probability of being cited among documents of similar topics and Pairwise-Link-LDA does not have this flexibility. In terms of CPU times, RTM is the fastest as it scales as $O(DK)$ followed by LDA+Regression. LMV and Pairwise-Link-LDA took much longer to converge as they scale as $O(D^2 K^2)$. However, subsampling can help to reduce the runtimes while producing equally good predictions. In examples where the data size is large, subsampling also helps to overcome memory constraints.

\begin{figure}[tb!]
	\centering
	\includegraphics[width=0.65\textwidth]{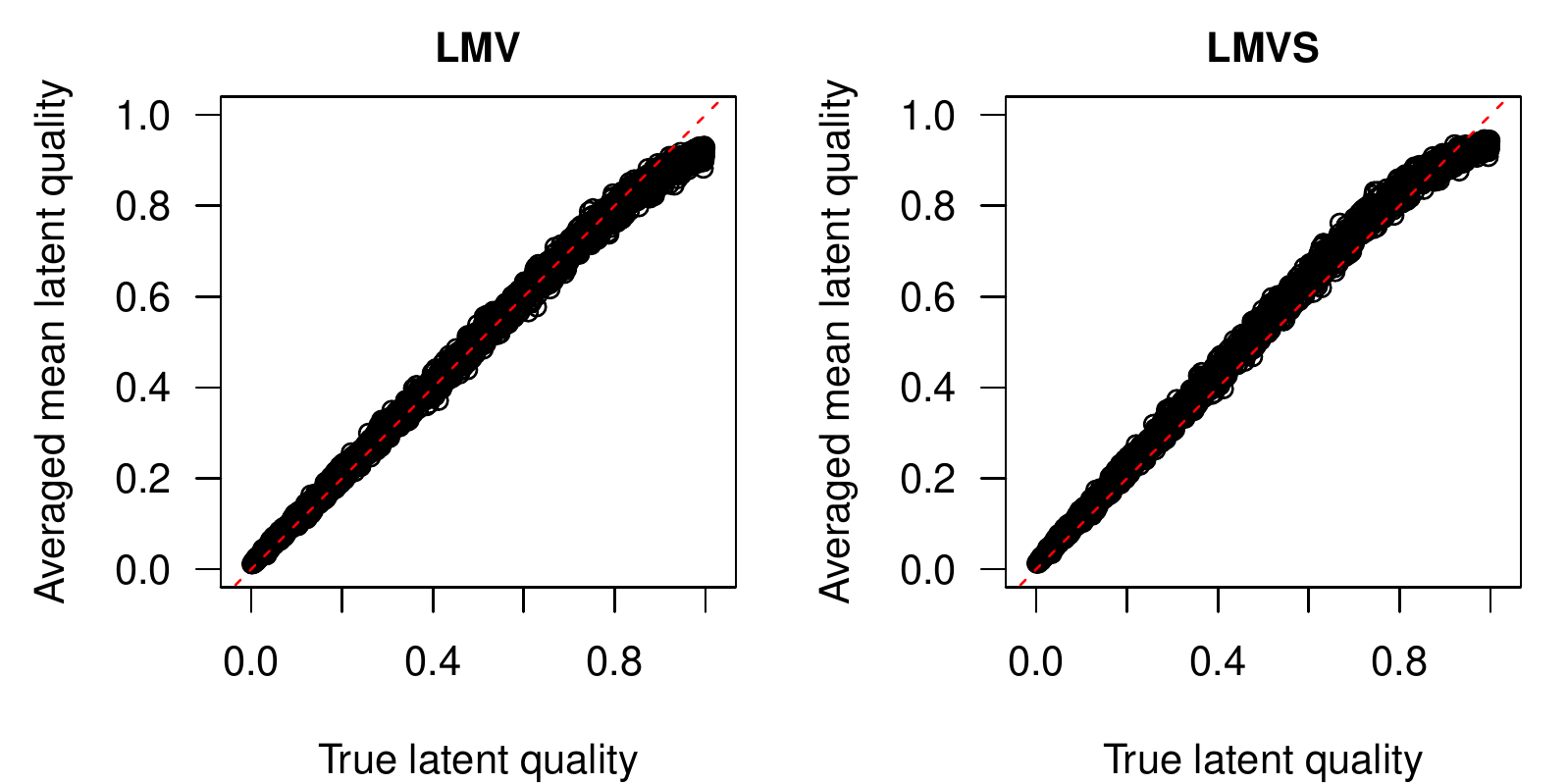}
	\caption {\label{simlq} Plots of posterior mean visibility (averaged over 20 datasets) against true visibility for Algorithms 1 and 2.}
\end{figure}

Next, we study the accuracy of Algorithms 1 and 2. Figure \ref{simlq} plots the posterior mean visibility averaged over 20 datasets against the true visibility. 
The estimated posterior mean visibility is close to the true visibility for both algorithms 1 and 2. However, there is a slight underestimation of visibilities that are very close to 1 in both Algorithms and a slight overestimation more generally in Algorithm 2. Table \ref{Best} shows the mean and standard deviation of the posterior mean estimates of elements in the blockmodel $B$ computed using LMV (Algorithms 1 and 2) and Pairwise-Link-LDA. 
\begin{table}[htb!]
	\caption{Mean and standard deviation (after $\pm$) of the posterior mean estimates of nonzero blockmodel elements over 20 simulated datasets.}
	\label{Best}
	\centering
	\small
	\begin{tabular}{lcccc} 
		\hline
		True value  &   LMV    &   LMVS &  PLLDA \\ \hline
		$B$[1,1]: 0.3   &  0.298 $\pm$ 0.005  &  0.279 $\pm$ 0.007  & 0.146 $\pm$ 0.004\\
		$B$[2,2]: 0.3   &  0.299  $\pm$ 0.004 & 0.278 $\pm$ 0.006   & 0.147 $\pm$ 0.003\\
		$B$[3,3]: 0.3   & 0.297   $\pm$  0.005 & 0.278 $\pm$ 0.007  & 0.147 $\pm$ 0.005\\
		$B$[4,4]: 0.2   &  0.200  $\pm$ 0.004  & 0.187 $\pm$ 0.004  &  0.098 $\pm$ 0.003\\
		$B$[5,5]: 0.2   & 0.203  $\pm$ 0.004  & 0.187 $\pm$ 0.006  &  0.099 $\pm$ 0.003\\
		$B$[6,6]: 0.2   & 0.201 $\pm$ 0.004  & 0.187 $\pm$ 0.003  &  0.097 $\pm$ 0.003\\
		$B$[2,5]: 0.05 &  0.049  $\pm$  0.001 &  0.049 $\pm$ 0.003  &  0.024 $\pm$ 0.001 \\
		$B$[3,4]: 0.05 &  0.048  $\pm$  0.001   &  0.049 $\pm$ 0.002 &  0.023 $\pm$ 0.001 \\ 
		$B$[4,1]: 0.04 & 0.039   $\pm$ 0.001 & 0.039 $\pm$ 0.002  &  0.019 $\pm$ 0.001\\    
		$B$[5,2]: 0.04 & 0.039   $\pm$ 0.001 & 0.039 $\pm$ 0.002 &  0.019 $\pm$ 0.001\\    
		$B$[1,6]: 0.03 &  0.030  $\pm$ 0.001 & 0.030 $\pm$  0.001 &  0.014 $\pm$ 0.001 \\
		$B$[6,3]: 0.03 & 0.029  $\pm$ 0.001 & 0.029 $\pm$ 0.001 & 0.014 $\pm$ 0.001\\
		$B$[1,3]: 0.02 &  0.020 $\pm$ 0.001   & 0.020 $\pm$ 0.001 & 0.010 $\pm$ 0.000\\
		$B$[4,6]: 0.02 & 0.020   $\pm$ 0.001 & 0.021 $\pm$ 0.001 &  0.010 $\pm$ 0.000\\
		$B$[6,1]: 0.02 & 0.020 $\pm$  0.001 & 0.020 $\pm$ 0.001 &  0.010 $\pm$ 0.001\\    \hline
	\end{tabular}
\end{table}
Only estimates of nonzero elements of $B$ are shown. For elements of $B$ which are zero, all estimates are less than 0.0022 across all simulated datasets for the LMV (Algorithms 1 and 2) and Pairwise-Link-LDA. The estimates from Algorithm 1 are very close to the true values in $B$. There is slight underestimation in Algorithm 2 and greater variation in results across different datasets due to the subsampling of documents and zeros. It is also interesting to note that estimates of the blockmodel in Pairwise-Link-LDA are much smaller or about half of the true values in $B$. As Pairwise-Link-LDA does not distinguish between documents of similar topics with different visibilities, and the visibilities were simulated from Uniform[0,1], there is an averaging effect where elements of $B$ are scaled by 0.5. This simulated example shows that the LMV has the potential to improve citation predictions in real-world scenarios. While running the LMV in batch mode is time consuming, subsampling can reduce the runtime and memory requirements while producing competitive estimations and predictions.

\end{document}